\documentclass[runningheads]{llncs}

\usepackage{eccv}

\usepackage{eccvabbrv}

\usepackage{graphicx}
\usepackage{booktabs}

\usepackage{enumitem}
\usepackage{multirow}

\newcounter{tableeqn}[table]
\renewcommand{\thetableeqn}{\thetable.\arabic{tableeqn}}
\newcounter{tablesubeqn}[tableeqn]

\definecolor{dartmouthgreen}{rgb}{0.05, 0.5, 0.06}
\newcommand{\blue}[1]{\textcolor{dartmouthgreen}{#1}}
\newcommand{\red}[1]{\textcolor{red}{#1}}
\definecolor{blue(ncs)}{rgb}{0.0, 0.53, 0.74}
\newcommand{\bkbone}[1]{\textcolor{blue(ncs)}{#1}}

\DeclareMathOperator*{\topmaxkappa}{TopMax_{\text{$\kappa$}}}

\usepackage[accsupp]{axessibility}  

\usepackage{hyperref}

\usepackage{orcidlink}

\usepackage{amsmath,amsfonts,bm}

\newcommand{\mbr}[1]{\mathbb{R}^{#1}}

\def\eqref#1{(\ref{#1})}

\def\1{\bm{1}}

\def\vh{{\bm{h}}}

\def\vs{{\bm{s}}}

\def\vx{{\bm{x}}}

\def\vz{{\bm{z}}}

\DeclareMathAlphabet{\mathsfit}{\encodingdefault}{\sfdefault}{m}{sl}
\SetMathAlphabet{\mathsfit}{bold}{\encodingdefault}{\sfdefault}{bx}{n}

\DeclareMathOperator*{\argmax}{arg\,max}
\DeclareMathOperator*{\argmin}{arg\,min}

\makeatletter
\DeclareRobustCommand\onedot{\futurelet\@let@token\bmv@onedotaux}
\def\bmv@onedotaux{\ifx\@let@token.\else.\null\fi\xspace}
\def\eg{\emph{e.g}\onedot} 
\def\ie{\emph{i.e}\onedot} 
 
\def\etc{\emph{etc}\onedot} \def\vs{\emph{vs}\onedot}
\def\wrt{w.r.t\onedot} 
\def\aka{a.k.a\onedot}

\makeatother

\begin{document}

\title{
Adaptive Multi-head Contrastive Learning
} 

\titlerunning{Adaptive Multi-head Contrastive Learning}




\author{Lei Wang\thanks{Corresponding author.}\inst{, 1,2}\orcidlink{0000-0002-8600-7099}\and Piotr Koniusz\inst{2,1}\orcidlink{0000-0002-6340-5289} \and Tom Gedeon\inst{3}\orcidlink{0000-0001-8356-4909} \and Liang Zheng\inst{1}\orcidlink{0000-0002-1464-9500}\\
}

\institute{$^{1}$Australian National University \;
$^{2}$Data61/CSIRO \;
$^{3}$Curtin University \\
\email{\{lei.w, liang.zheng\}@anu.edu.au}, \email{piotr.koniusz@data61.csiro.au}, \email{tom.gedeon@curtin.edu.au}}

\maketitle

\begin{abstract}
  In contrastive learning, two views of an original image, generated by different augmentations, are considered a positive pair, and their similarity is required to be high. Similarly, two views of distinct images form a negative pair, with encouraged low similarity. Typically, a single similarity measure, provided by a lone projection head, evaluates positive and negative sample pairs. However, due to diverse augmentation strategies and varying intra-sample similarity, views from the same image may not always be similar. Additionally, owing to inter-sample similarity, views from different images may be more akin than those from the same image. Consequently, enforcing high similarity for positive pairs and low similarity for negative pairs may be unattainable, and in some cases, such enforcement could detrimentally impact performance. To address this challenge, we propose using multiple projection heads, each producing a distinct set of features. Our pre-training loss function emerges from a solution to the maximum likelihood estimation over head-wise posterior distributions of positive samples given observations. This loss incorporates the similarity measure over positive and negative pairs, each re-weighted by an individual adaptive temperature, regulated to prevent ill solutions. Our approach, Adaptive Multi-Head Contrastive Learning (AMCL), can be applied to and experimentally enhances several popular contrastive learning methods such as SimCLR, MoCo, and Barlow Twins. The improvement remains consistent across various backbones and linear probing epochs, and becomes more significant when employing multiple augmentation methods. Code is \href{https://github.com/LeiWangR/cl}{here}.
\end{abstract}

\section{Introduction}
\label{sec:intro}

\begin{figure*}[tb]
\begin{subfigure}[b]{0.5\linewidth}
\centering
\includegraphics[trim=0.5cm 4cm 0cm 1cm, width=\textwidth]{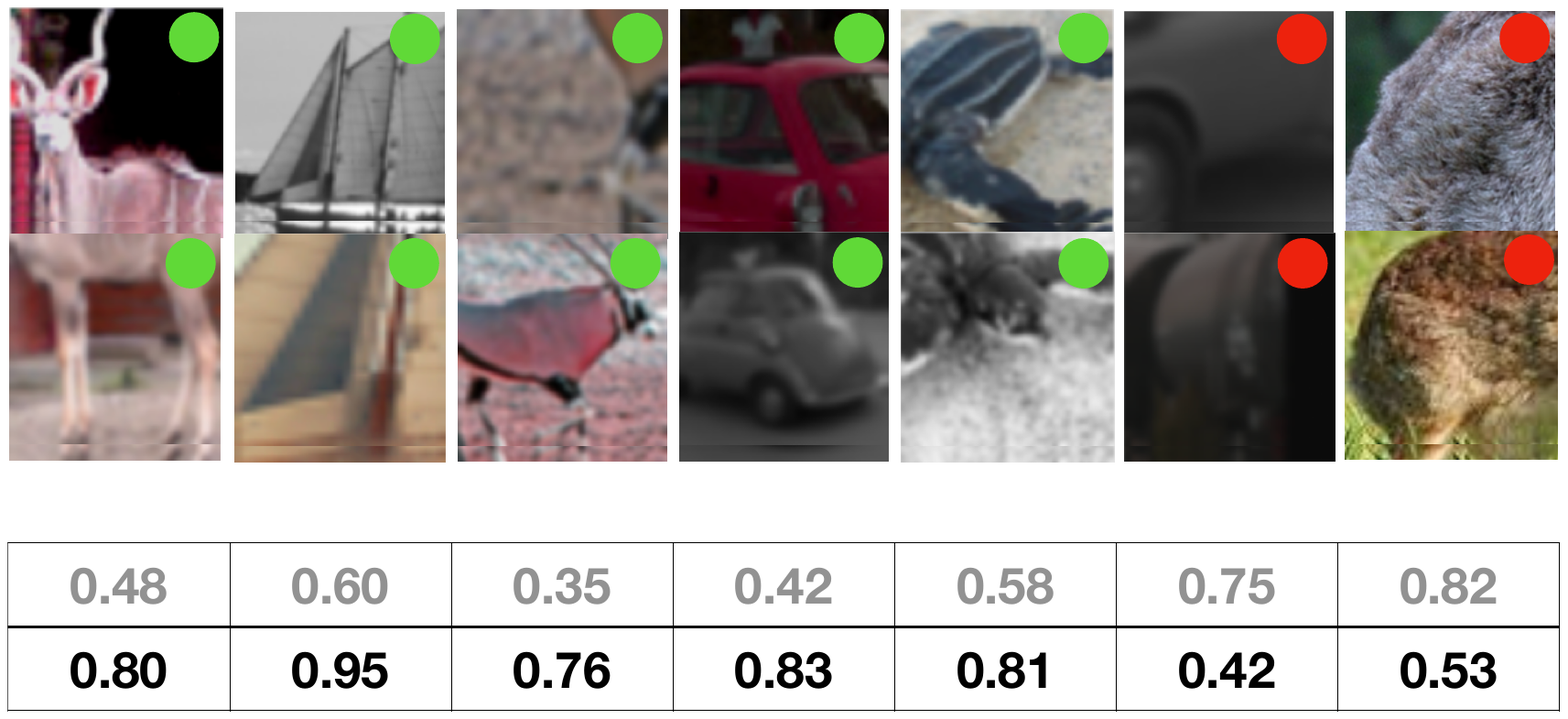}
\captionsetup[subfigure]{font=small}
\caption{Examples of augmented image pairs (STL-10).\label{fig:stl10-aug}}
\end{subfigure}\hfill
\begin{minipage}[b]{0.5\linewidth}
\centering
\begin{subfigure}{0.33\linewidth}
  \includegraphics[width=\linewidth]{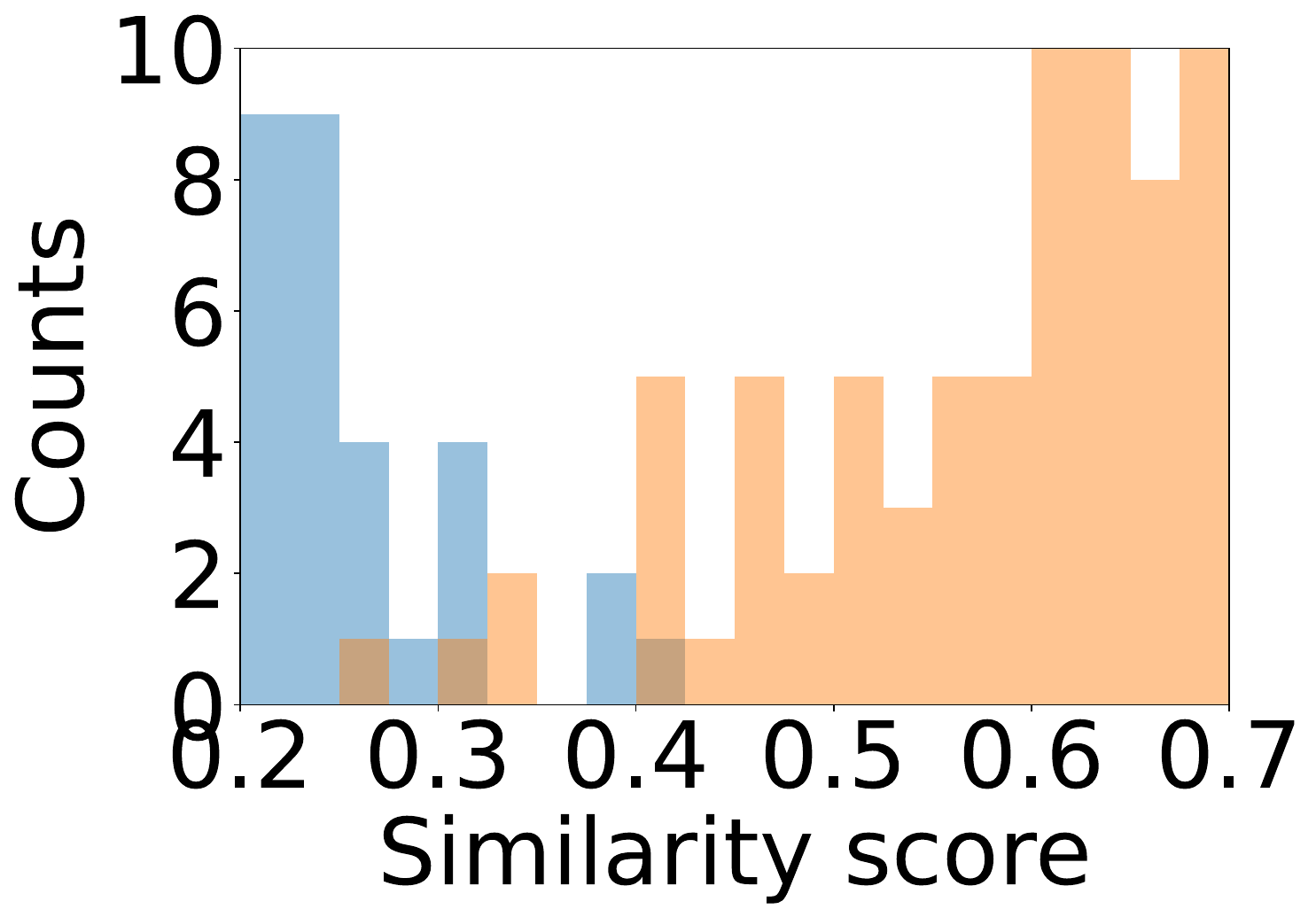}
  \caption{$\!\!$1 head,1 aug.}\label{fig:1-aug1}
\end{subfigure}\hfill
\begin{subfigure}{0.33\linewidth}
  \includegraphics[width=\linewidth]{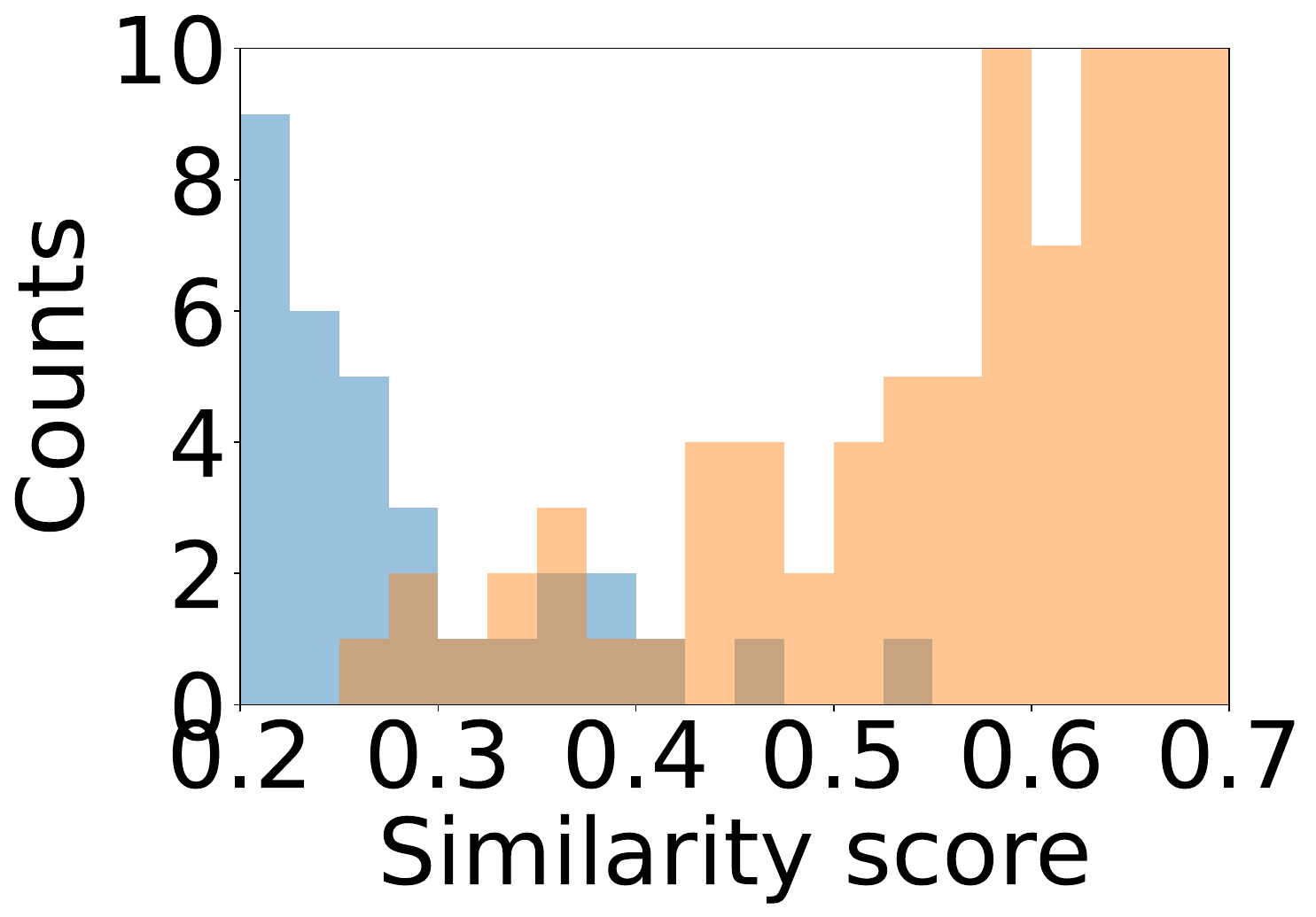}
  \caption{$\!\!$1 head,3 aug.}\label{fig:2-aug3}
\end{subfigure}\hfill
\begin{subfigure}{0.33\linewidth}
  \includegraphics[width=\linewidth]{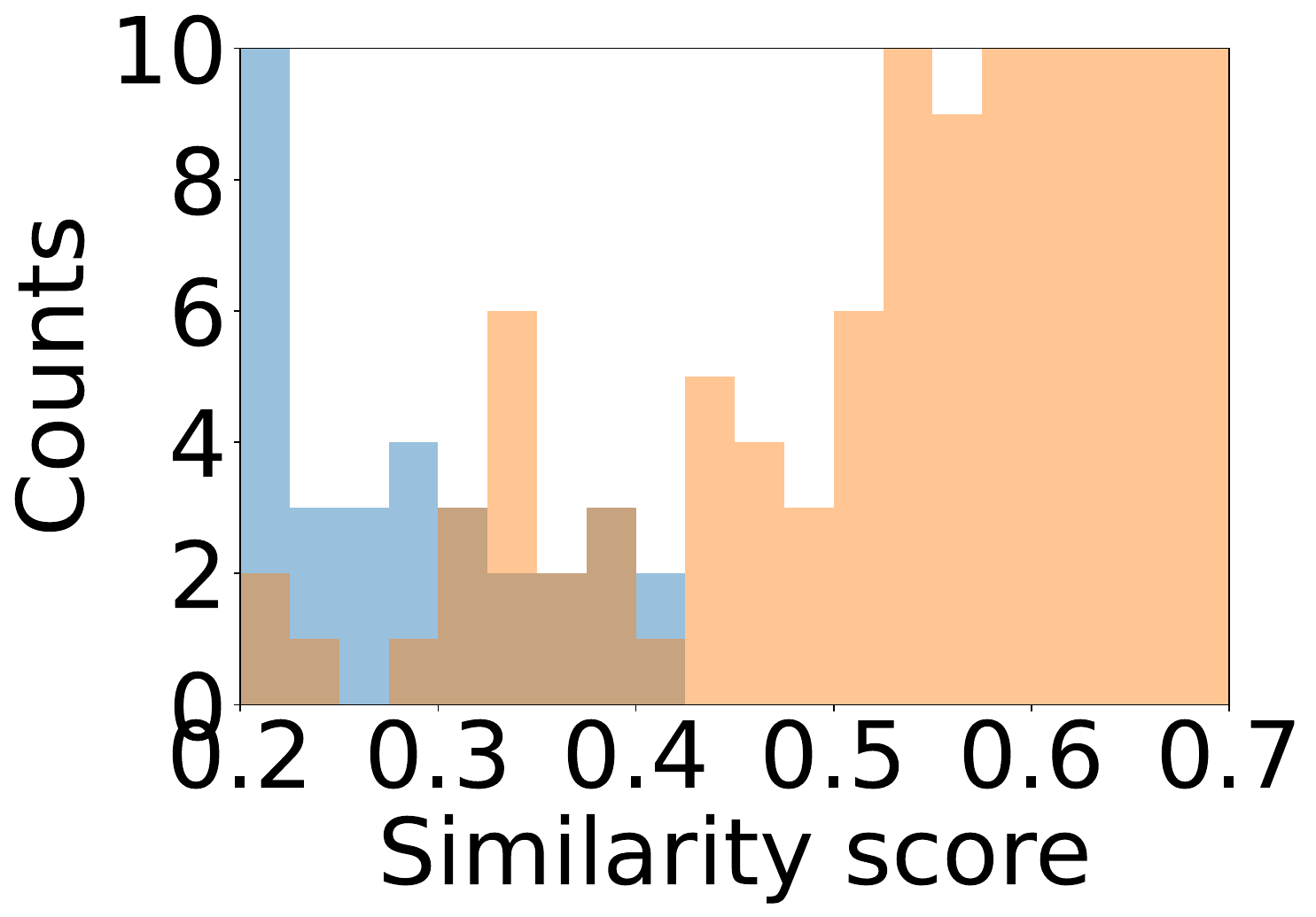}
  \caption{$\!\!$1 head,5 aug.}\label{fig:3-aug5}
\end{subfigure}
\begin{subfigure}{0.33\linewidth}
  \includegraphics[width=\linewidth]{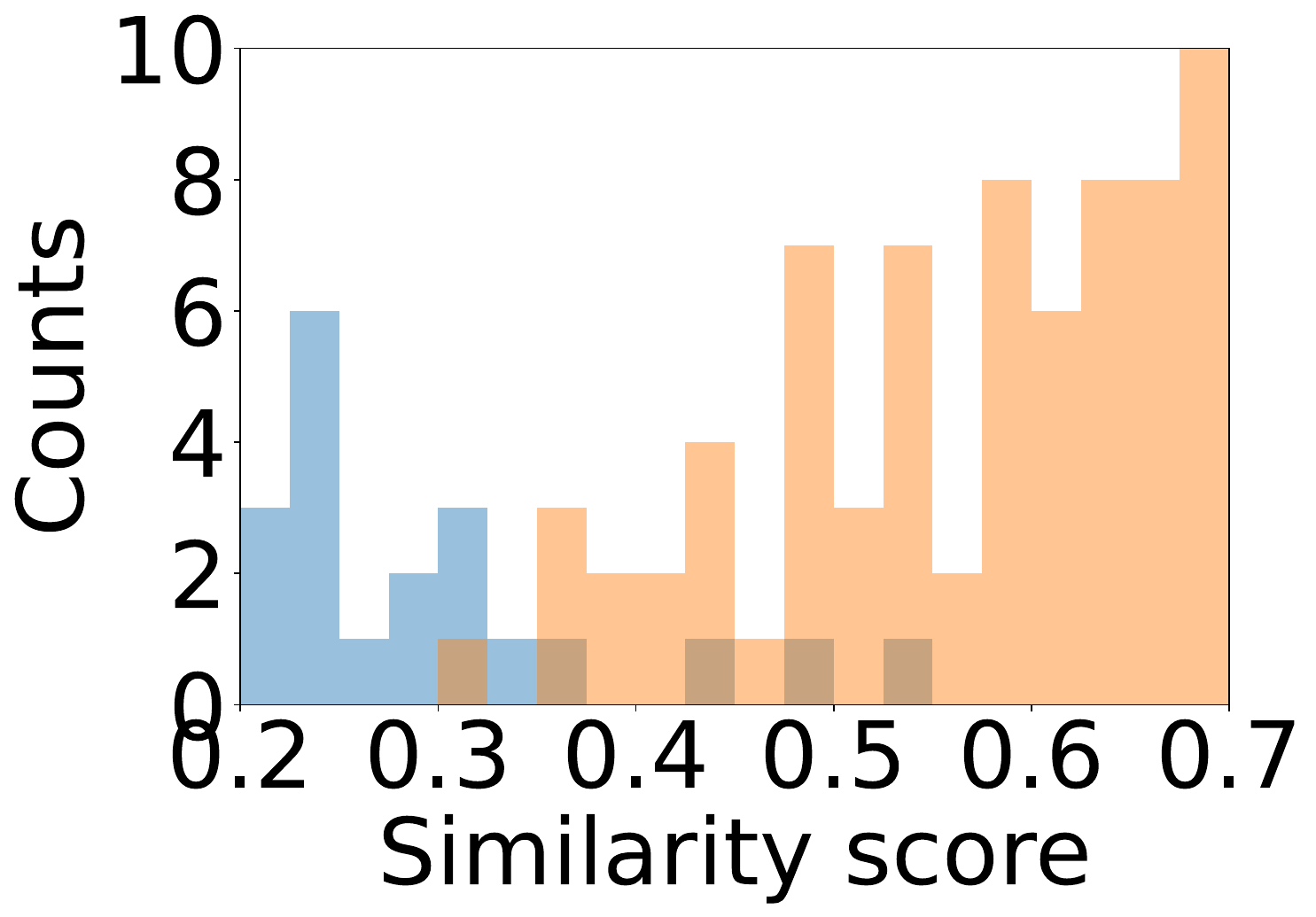}
  \caption{Ours, 1 aug.}\label{fig:4-aug1}
\end{subfigure}\hfill
\begin{subfigure}{0.33\linewidth}
  \includegraphics[width=\linewidth]{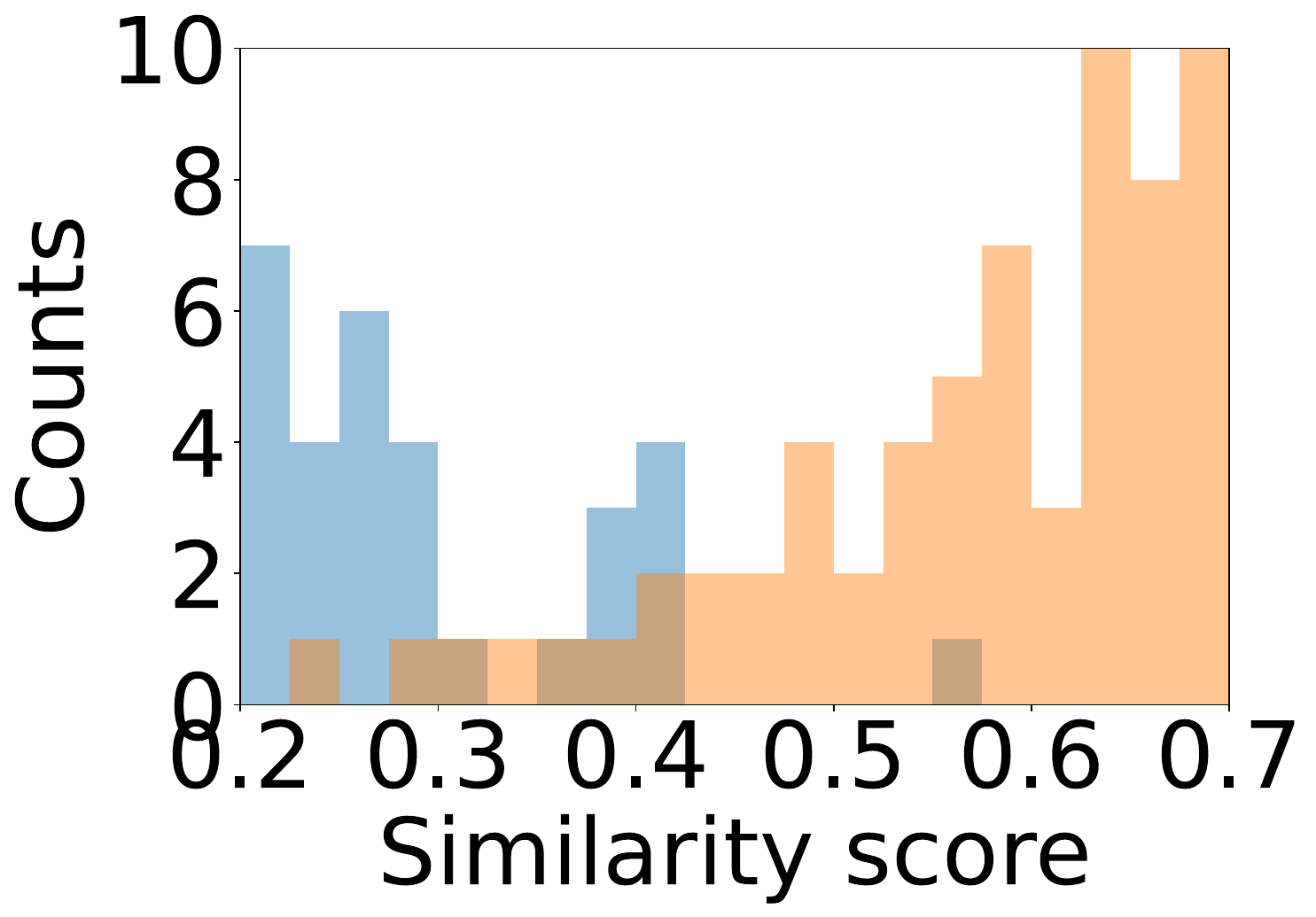}
  \caption{Ours, 3 aug.}\label{fig:5-aug3}
\end{subfigure}\hfill
\begin{subfigure}{0.33\linewidth}
  \includegraphics[width=\linewidth]{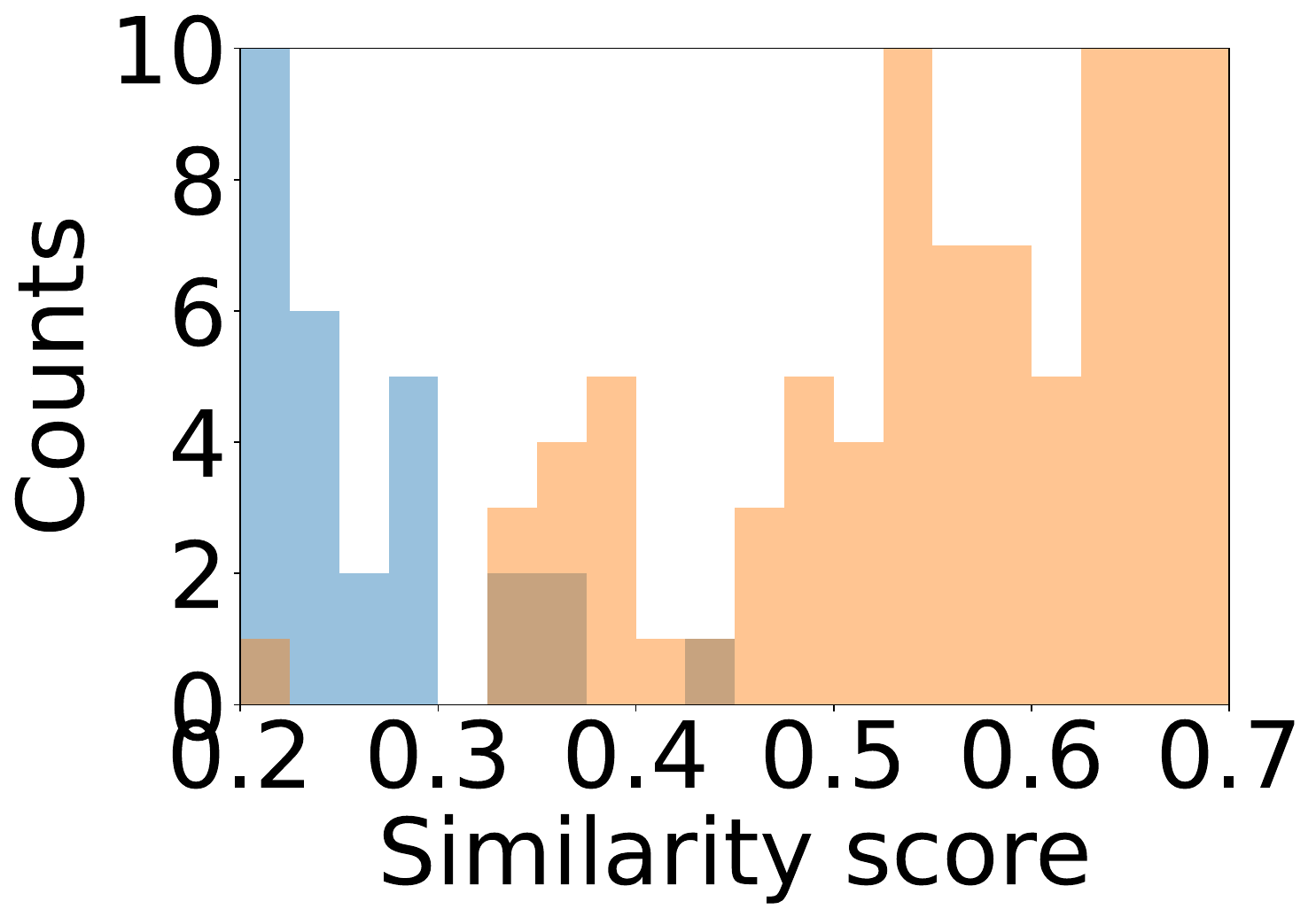}
  \caption{Ours, 5 aug.}\label{fig:6-aug5}
\end{subfigure}
\end{minipage}
\vspace{-0.2cm}
\caption{In (a), each column denotes positive (green dots) or negative instances (red dots) with corresponding similarity measures. Additional augmentations can cause positive samples to appear dissimilar and occasionally make negative samples seem similar. The table in (a) shows the original similarity measure (in gray) and the similarity scores from our method (in black).(b)-(d): for traditional contrastive learning methods, when increasing the number of augmentations from 1 to 5, similarities of more positive pairs drop below 0.5, causing more significant overlapping regions between histograms of positive (orange) and negative (blue) pairs.
In comparison, our multi-head approach (e)-(g) 
yields better separation of positive and negative sample pairs as more augmentation types are used, \eg, (g) \vs (d).} 
\label{fig:motivation}
\vspace{-0.5cm}
\end{figure*}

Contrastive learning is an important line of work in self-supervised learning (SSL) which offers a promising path to leveraging large quantities of unlabeled data. Its main idea is to encourage two views of the same image (positive pair) to have similar embeddings and thus a high similarity, and those of different images (negative pair) to have a low similarity. As such, the similarity measure is an important component influencing representation learning. 

In literature, multiple augmentations are usually used to create a view of an image. For example, rotation, scaling, translation, and flipping are used in SimCLR~\cite{chen2020simple} and MoCo~\cite{he2020momentum}. However, the use of multiple augmentations make positive pairs often look dissimilar and  negative pairs occasionally similar: examples are presented in Fig.~\ref{fig:motivation}(a). Therefore, there exists non-negligible diversity in the distribution of similarity of image pairs. As shown in Fig.~\ref{fig:motivation}(b)-(d), when the number of augmentations increases from 1, 3 to 5, the similarity distributions of positive and negative pairs of the SimCLR method become more complex, \emph{e.g.}, the similarity of more positive pairs drops below 0.5; thus, we observe increasingly significant overlapping regions, indicating compromised similarity learning.

We identify two limitations of existing methods which prevent them from addressing the above-mentioned problem. \textit{First}, existing methods usually use a single feature projection head and a single similarity measure \cite{chen2020simple,he2020momentum,chen2021exploring}. While this head is supervised by standard metric loss such as contrastive loss, a single projection has a single mode of image characterization which would be insufficient to describe the diverse image content caused by multiple augmentations. A consequence is that positive pairs sometimes have low similarity scores. 
\textit{Second}, existing methods usually use a global temperature to scale similarity, which, after careful tuning, is shown to improve feature alignment and uniformity \cite{wang2020understanding}. 
 Under this scheme, the same scaling is applied to all the pairs, which does not alleviate overlapping exhibited in Fig. \ref{fig:motivation}(d).

This paper aims to address the diversity issue caused by multiple augmentations while considering the limitations in existing practice. We propose adaptive multi-head contrastive learning (AMCL): it better captures the diverse image content and gives similarity scores that better separate positive and negative pairs. In a nutshell, instead of having a single MLP and cosine similarity, AMCL uses multiple repetitive MLPs and cosine similarity measures before loss computation. Within AMCL, we design an adaptive temperature which depends on both the projection head and the similarity of the current pair. We show that the idea of multiple projection heads and adaptive temperature can be applied to popular contrastive learning frameworks and yields consistent improvements. Aligned with our motivation, we observe more pronounced enhancements with 4-5 augmentation types compared to 1-2 types.

On the theoretical side, we derive training objective function of AMCL based on maximum likelihood estimation (MLE). We show this objective function can be reduced to many existing contrastive learning methods, that its regularization term has interesting physical meaning, and that with it we are now able to connect temperature to uncertainty. We summarize the main points below. 

\renewcommand{\labelenumi}{\roman{enumi}.}
\begin{enumerate}[leftmargin=0.6cm] 
\item We propose adaptive multi-head contrastive learning (AMCL) which tackles intra- and inter-sample similarity and an adaptive temperature mechanism re-weighting each similarity pair. 
\item We derive the objective function for AMCL as solution to the maximum likelihood estimation. We also discuss its mathematical insights including connecting temperature to uncertainty. 

\item Our system consistently improves the performance of a few popular constrastive learning frameworks, backbones, loss functions, and combinations of augmentation types, and is shown to be particularly useful under more augmentation types.
\end{enumerate}

\section{Related Work}
\label{sec:related}

\textbf{Self-supervised learning} (SSL) has been a driving force behind unsupervised learning in computer vision~\cite{lei_mm_21} and natural language processing (NLP)~\cite{balestriero2023cookbook}. Its methods can be grouped into 4 broad families: deep metric learning~\cite{chen2020simple, dwibedi2021little, du2021self},  
self-distillation~\cite{grill2020bootstrap,chen2021exploring,caron2021emerging,he2020momentum, zhou2021ibot, Koohpayegani_2021_ICCV, oquab2023dinov2},
canonical correlation analysis~\cite{da6385d2-9c65-3860-bbcd-b821fdff69ff, zbontar2021barlow,caron2020unsupervised,bardes2022vicreg}, and masked image modeling (MIM)~\cite{bao2022beit, he2022masked,xie2022simmim}.
Since contrastive learning and MIM can complement each other~\cite{park2023what},  
recent works have adopted a fusion of both to improve representation quality and transfer performance over its traditional MIM approaches~\cite{huang2022contrastive, mishra2022simple, wei2022contrastive, jiang2023layer}.

\noindent\textbf{Metric learning} is related to self-supervised learning~\cite{balestriero2023cookbook}. Commonly used similarity measurements include the triplet loss~\cite{hoffer2015deep}, cross-entropy loss~\cite{zhang2018generalized}, and contrastive loss~\cite{khosla2020supervised}. A contemporary work is multi-similarity learning~\cite{mu2023multi}, where different attribute labels of an image are used in each level of learning. Different from~\cite{mu2023multi}, our method is self-supervised, discusses the adaptive temperature as a useful add-on, and derives interesting mathematical insights.

\noindent{\bf Uncertainty learning} has been studied extensively~\cite{abdar2021review, gawlikowski2023survey}. \cite{tao2019deep} uses multiple network copies trained with different parameter initializations to find various local minima. 
Bayesian neural networks~\cite{goan2020bayesian} and Monte Carlo dropout~\cite{gal2016dropout} handle uncertainty by design, where dropout layers 
are equivalent to sampling weights from a posterior distribution over model parameters. 
A more principled way is to capture aleatoric uncertainty \cite{uncertainty1,uncertainty5,uncertainty4} of Euclidean distance or cosine similarity, \eg, heteroscedastic aleatoric uncertainty  (observation noise may vary with each pair of samples). 
To this end, we model the maximum likelihood estimation over head-wise posterior distributions of positive samples given observations. This is a form of m-estimator  \cite{Huber.Wiley} whose  log-likelihood employs Normal distributions \textit{\aka} Welsch functions by the uncertainty estimation community.

\section{SSL Frameworks: A revisit}

\begin{table*}[tb]
\centering
\caption{Standard contrastive learning methods and their loss functions.}
\vspace{-0.2cm}
\label{tab:loss_functions}
\resizebox{\textwidth}{!}{\begin{tabular}{l  l l c } 
\toprule
Method & Loss name & $\qquad\qquad\qquad\qquad$Loss function \\
\midrule
SimCLR, MoCo & NT-Xent & $\ell_\text{NT-Xent}\;\;\;=\! -\text{log}\;
\frac{\text{exp}(\text{sim}(\vz_i, \vz^+_i)/\tau)}{\sum_{n=1}^{N}\text{exp}(\text{sim}(\vz_i, \vz^-_{in})/\tau)}
$ & $\!\!\!\!$\refstepcounter{tableeqn} (\thetableeqn)\label{eq:ntxent} \\
SimSiam & $\!\!$Negative cos. &  $\ell_\text{SymNegCos}
    \!=\!- \frac{1}{2}  \text{sim}\big(\vz_i, [\vh^+_i]_\text{sg}\big) - \frac{1}{2} \text{sim}\big(\vz^+_i, [\vh_i]_\text{sg}\big) $ & $\!\!\!\!$\refstepcounter{tableeqn} (\thetableeqn)\label{eq:simsiam} \\
Barlow Twins & $\!\!$Cross-corr. & $\ell_{\text{Cross-Corr}} \;=\! \sum_{l=1}^{d'}(1-\mathcal{C}_{ll})^2 + \lambda \sum_{l=1}^{d'} \sum_{m \neq l}^{d'} \mathcal{C}_{lm}^2$ & $\!\!\!\!$\refstepcounter{tableeqn} (\thetableeqn)\label{eq:bt} \\
LGP, CAN & $\!\!$InfoNCE & $\ell_\text{InfoNCE}\;\;\;=\! -\text{log}\;
\frac{\text{exp}(\text{sim}(\vz_i, \vz^+_i)/\tau)}{{\text{exp}(\text{sim}(\vz_i, \vz^+_i)/\tau)}+\sum_{n=1}^{N}\text{exp}(\text{sim}(\vz_i, \vz^-_{in})/\tau)}
$ & $\!\!\!\!$\refstepcounter{tableeqn} (\thetableeqn)\label{eq:infonce} \\
\bottomrule
\end{tabular}}
\vspace{-0.5cm}
\end{table*}

\textbf{Notations}. A common contrastive learning framework typically consists of a data augmentation module, a base encoder $f(\cdot)$, a projection head $g(\cdot)$, and a loss function $\ell_\text{Contrast}(\cdot)$. 
Stochastic data augmentation transforms a given sample randomly, resulting in two views of the same sample denoted $\vx_i$ and $\vx^+_i$, which are considered as a positive pair consisting of an anchor and positive sample, respectively. Their visual representations are denoted as $\vh_i\!=\!f(\vx_i)\in\!\mbr{d}$ and $\vh^+_i\!=\!f(\vx^+_i)\in\!\mbr{d}$, where $d$ is feature dimension. 
The projection head $g(\cdot)$ maps these $d$-dim vectors to $d'$-dim vectors $\vz_i\!=\!g(\vh_i)\in\!\mbr{d'}$ and $\vz^+_i\!=\!g(\vh^+_i)\in\!\mbr{d'}$, to which the contrastive learning loss is applied. Normally one single multi-layer perceptron (MLP) is used for projection.  
By analogy,  
negative samples for anchor $\vx_i$ are denoted by $\vx^-_{in}$ ($n\!=\!1, \cdots, N$, and $N$ is the total number of negative samples per anchor), and their features and projection head outputs are $\vh^-_{in}\!=\!f(\vx^-_{in})$ and $\vz^-_{in}\!=\!g(\vh^-_{in})$, respectively.
 
\noindent\textbf{Loss functions}. The contrastive loss function 
typically tries to align the anchors with their positive samples, and enlarge the distance between the anchors and their negative samples. Loss functions of some popular SSL methods are summarized in Table \ref{tab:loss_functions}. 
Methods such as 
SimCLR  and MoCo use the NT-Xent loss, 
given in Eq. \eqref{eq:ntxent}. 
NT-Xent is  very similar to the InfoNCE loss but differs by the normalizaton step.
Function $\text{sim}(\cdot, \cdot)$ in equations of Table \ref{tab:loss_functions} represents the cosine similarity. SimSiam uses the negative cosine similarity loss in Eq. \eqref{eq:simsiam}, 
where $[\cdot]_\text{sg}$ is the stop-gradient operation.  
Barlow Twins in Eq. \eqref{eq:bt} takes a different approach by utilizing the cross-correlation loss to decorrelate the channels of both views. 
In Eq. \eqref{eq:bt},  
$\lambda\geq 0$ is a hyperparameter that controls the strength of decorrelation. $\mathcal{C}$ is the cross-correlation matrix computed between the outputs of two identical networks along the batch dimension, \ie,
$\mathcal{C}_{lm}=\sum_{n=1}^N z_{ln}\,z^+_{mn}$~\cite{zbontar2021barlow}.

Differing from contrastive learning, masked image modeling (MIM) learns to reconstruct a corrupted image where some parts of the image or feature map are masked out. 
As demonstrated in~\cite{park2023what}, contrastive learning and MIM are complementary strategies. Thus,  recent works, 
LGP~\cite{jiang2023layer} and CAN~\cite{mishra2022simple}, combine the MIM loss 
and the InfoNCE loss in Eq. \eqref{eq:infonce}. Kindly notice our innovations apply to the contrastive losses rather than MIM.

\section{Approach}

\begin{figure*}[tb]
\centering
\captionsetup[subfigure]{font=small}
\begin{subfigure}{0.27\textwidth}
  \includegraphics[trim=3cm 20cm 25cm 0cm, width=\linewidth]{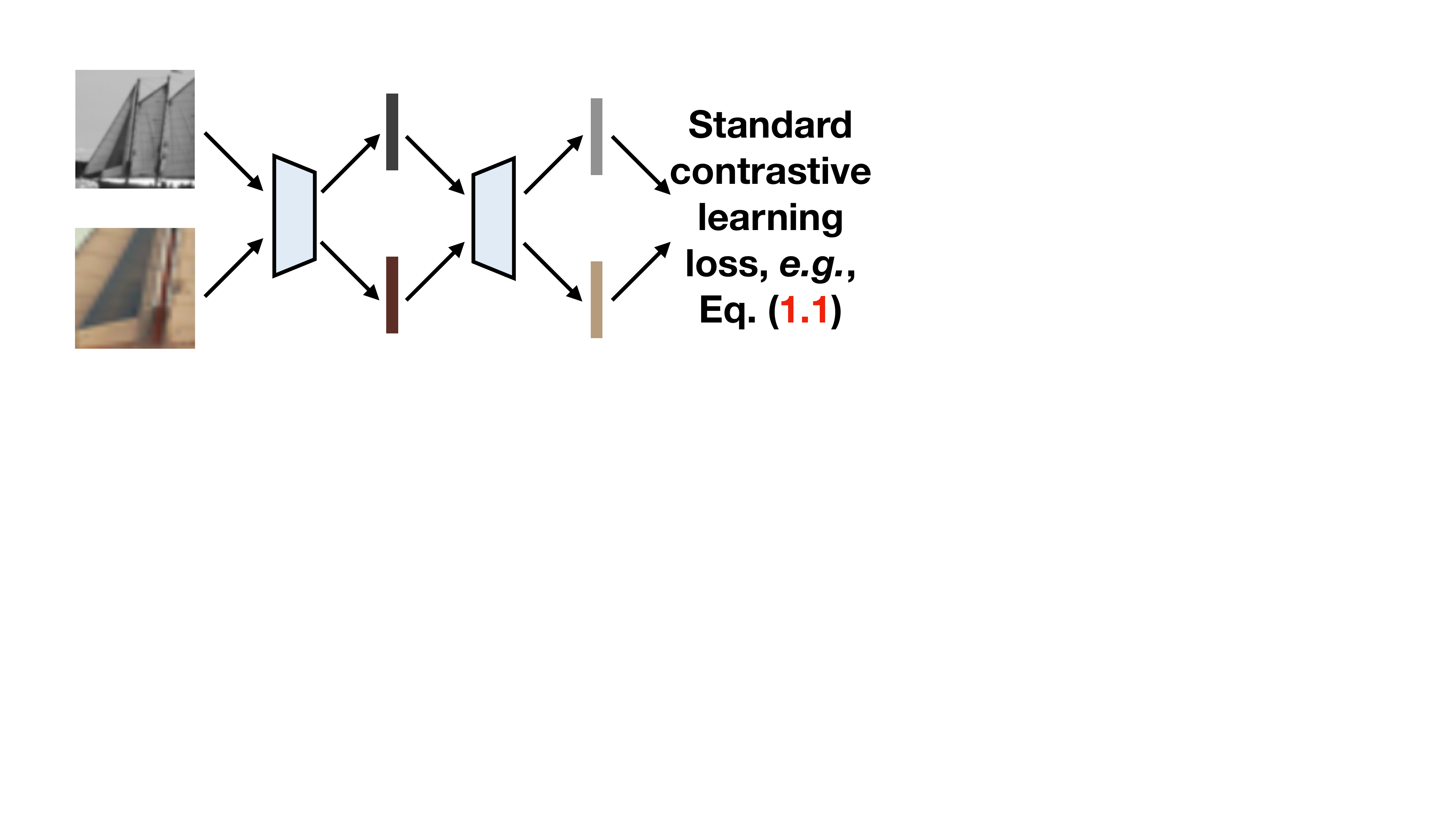}
  \captionsetup[subfigure]{font=small}
  \caption{Single-head, constant temperature.}
\end{subfigure}\hfill
\begin{subfigure}{0.34\textwidth}
  \includegraphics[trim=3cm 20cm 14cm 0cm,width=\linewidth]{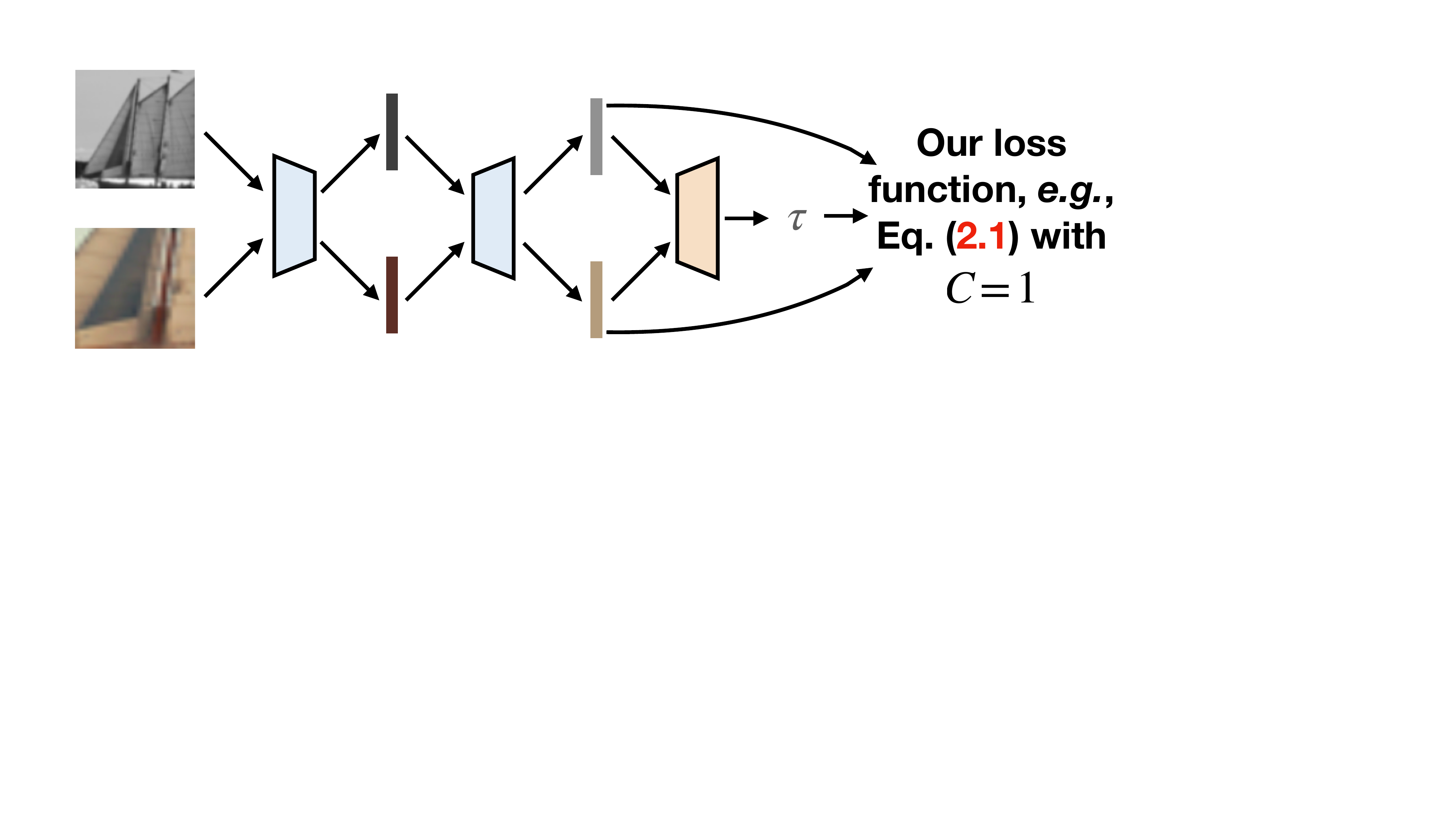}
  \caption{Single-head, adaptive temperature}
\end{subfigure}\hfill
\begin{subfigure}{0.34\textwidth}
  \includegraphics[trim=3cm 20cm 14cm 0cm,width=\linewidth]{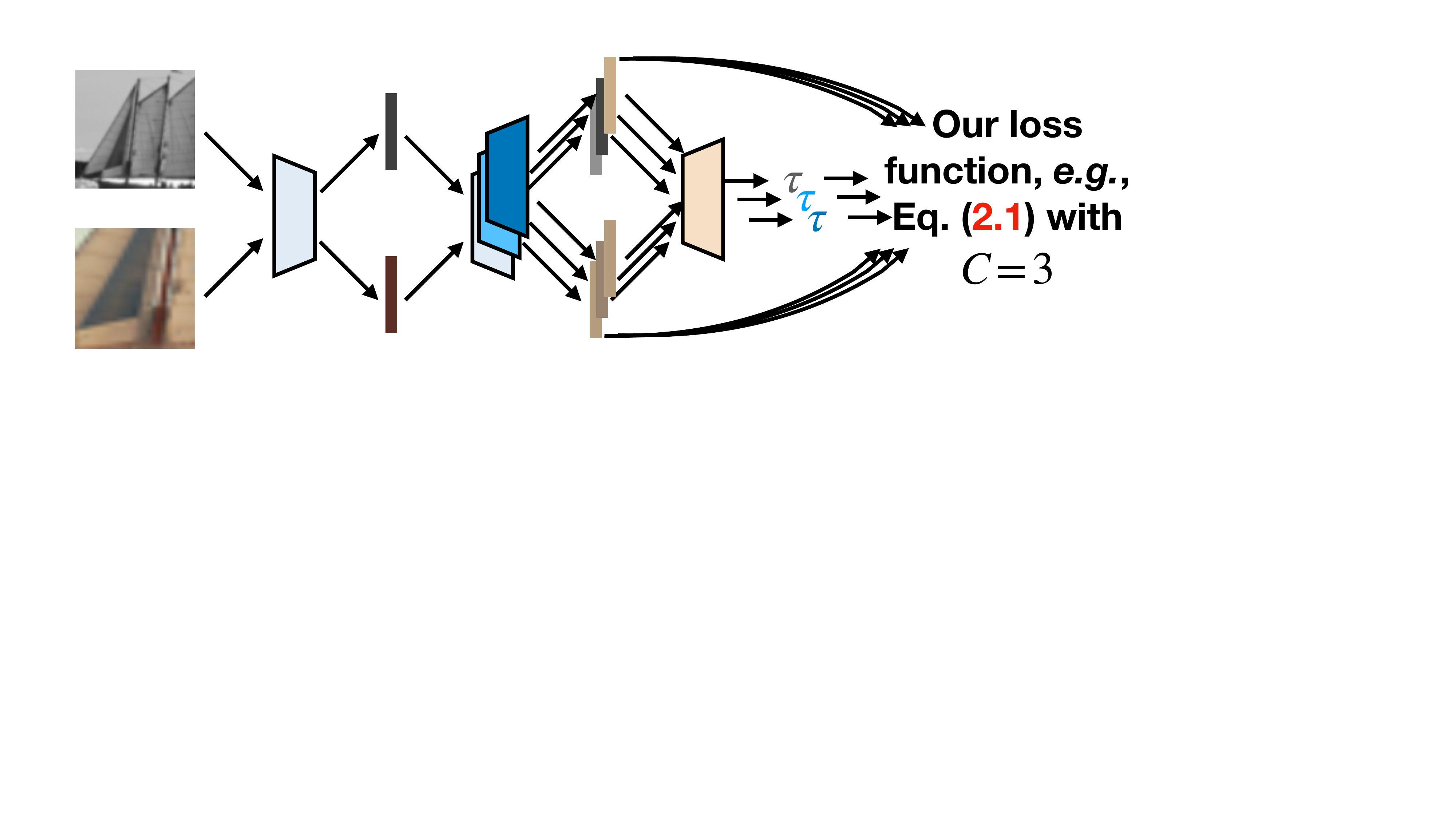}
  \caption{Multi-head, adaptive temperature.}
\end{subfigure}
\vspace{-0.2cm}
\caption{A comparison of (a) the standard constant-temperature, single-head approach and (b)–(c) our adaptive temperature, single- and multi-head approaches. 
In each subfigure, the first light blue trapezoid represents the base encoder, the second light blue trapezoid signifies the MLP projection head, and the third light orange trapezoid denotes the shared MLP layer for learning the temperature parameters. In (c), the projection head is replicated $C$ times to capture diverse image content. For better visualization, we set $C\!=\!3$ for simplicity. The architecture of the MLP projection head remains unchanged; however, the weights are learned independently.
For a given image pair, each projection head produces a pair of feature vectors, which are later used for learning the pair-adaptive, head-wise temperature. The learned temperatures, along with the projected features, are seamlessly incorporated into our Adaptive Multi-head Contrastive Learning (AMCL) loss function (as presented in Table~\ref{tab:loss_functions2}). 
}
\label{Fig:mainpipe}
\vspace{-0.5cm}
\end{figure*}

In this section, we present our Adaptive Multi-head Contrastive Learning (AMCL) framework, starting with learnable pair-adaptive temperatures and introducing AMCL loss functions. We then provide a comprehensive discussion, offering deeper insights into our framework.

\subsection{Learnable Pair-adaptive Temperature}

Existing contrastive learning frameworks generally adopt a constant temperature to scale similarity (see Fig.~\ref{Fig:mainpipe} (a)). The temperature parameter is painstakingly tuned and may lead to sub-optimal performance. For a given image pair (either positive or negative), the temperature parameter should be adaptive to multiple augmentations for a more robust and better similarity measure after temperature scaling. Furthermore, distinct temperature parameters should be applied to positive and negative pairs, considering the image content and emphasizing unique focal points within the images. We introduce learnable adaptive positive and negative temperatures for effectively addressing varying inter- and intra-sample similarity during the learning process.

For a given image pair feature representations after MLP projection head (either positive pair $\vz_i$, $\vz^+_i$ or negative pair $\vz_i$, $\{\vz_{in}\}_{n=1}^N$), our loss function takes the following general form:
\setcounter{equation}{2}
\begin{align}
\label{eq:ours}
&\ell^\dagger = \ell_\text{Contrast}\big(\vz_i, \vz^+_i, \{\vz_{in}\}_{n=1}^N\big)\!\!+\!\!\beta\Omega\big(\tau^{+}_i)\!\!-\!\!\beta\Omega\big(\{\tau^{-}_{in}\}_{n=1}^N \big),\\
& \qquad\text{ where }\quad \tau^{+}_i=\sigma\big(\langle\phi(\vz_i), \phi(\vz^+_i) \rangle\big) \text{ and } \tau^{-}_{in}=\sigma\big(\langle\phi(\vz_i), \phi(\vz^-_{in}) \rangle\big).\nonumber
\end{align}
In the above equation, $\tau^{+}_i$ and $\tau^{-}_{in}$ denote the learnable adaptive positive and negative temperatures, respectively. 
$\phi(\cdot):\mbr{d'}\rightarrow\mbr{d'}$ is an MLP layer\footnote{This MLP layer is separate from the MLP projection heads.} shared among all heads. 
$\langle \cdot, \cdot\rangle$ represents the dot product.
Sigmoid function $\sigma(r)=\frac{\iota}{1+\exp(r)}\!+\!\eta$ controls the lower and upper limits of the temperature, where $\iota$ and $\eta$ are hyperparameters. $\beta\geq 0$ controls the temperature regularization imposed by $\Omega(\cdot)$. The regularization is written as:
\begin{equation}
\Omega(\tau)=(d'/2)\log(\tau)+1/\tau,
\end{equation}
which encourages temperature $\tau$ to move towards $\tau=2/d'$. The derivation of the regularization term is detailed in Appendix~\ref{sec:deriv}.

In Fig.~\ref{Fig:mainpipe} (b), we illustrate the application of our learnable temperature to a standard (single MLP projection head \aka single-head) contrastive learning framework. Note that our learnable temperature is pair-adaptive.

\subsection{Multiple Projections}
Typical SSL methods incorporate a projection head $g(\cdot)$, often consisting of a 2- or 3-layer MLP with ReLU activation. 
This projection head has proven to be highly beneficial 
as removing final layers of a pre-trained deep neural network helps mitigate overfitting to the training task and helps learning downstream tasks better~\cite{balestriero2023cookbook}. 
As shown in Fig~\ref{fig:motivation}, a single projection (single-head) has a single mode of image characterization which would be insufficient to describe the diverse image content caused by multiple augmentations. Therefore, the application of adaptive temperature to a single-head approach (as depicted in Fig.~\ref{Fig:mainpipe} (b)) presents several challenges: (i) increased variability in positive pairs, leading to inefficiencies in learning diverse representations; (ii) an inability to address issues when different parts of the image may require varying attention or when handling complex image content; and (iii) a lack of robustness and adaptability in managing different transformations, augmentations, or input variations.

Inspired by~\cite{tao2019deep} in enhancing model performance, robustness, and generalization, we propose to apply $C$ MLP projection heads (while keeping the architecture unchanged), denoted as $g^1(\cdot), \ldots, g^C(\cdot)$.
The goal is to capture complementary aspects of similarity between views (see Fig.~\ref{Fig:mainpipe}(c)). Moreover, each given image pair benefits from a pair-adaptive, head-wise temperature, contributing to a more robust and refined similarity measure. 

\subsection{AMCL Loss}

\begin{table*}[tb]
\centering
\caption{Loss functions for applying AMCL to widely used contrastive learning frameworks involve introducing $C$ heads and a regularization term (highlighted in green).
}
\vspace{-0.2cm}
\label{tab:loss_functions2}
\resizebox{\textwidth}{!}{\begin{tabular}{l  l  l c} 
\toprule
Method &  $\qquad\qquad\qquad\qquad$Loss function & $\!\!\!\!\!\!\!\!$\blue{Regularization}\\
\midrule
{SimCLR, MoCo}  &  
{$\ell^{\ddagger}_\text{NT-Xent}\;\;\,=\!\sum\limits_{c=1}^C\!\Big(\! \!-\frac{1}{\tau^{c+}_i}\text{sim}(\vz^c_i, \vz^{c+}_i)\!+\!\frac{1}{\tau^{c-}_{in^*}}\max\limits_{n=1,\cdots,N}\text{sim}(\vz^c_i, \vz^{c-}_{in})\Big.$} & {$\Big.\!\!\blue{+\beta\Omega(\tau^{c+}_i)\!-\!\beta\Omega(\tau^{c-}_{in^*})}\Big)$} 
& $\!\!\!\!$\refstepcounter{tableeqn} (\thetableeqn)\label{eq:ntxent2}$\!\!$ \\
{SimSiam} &   {$\ell^{\ddagger}_\text{SymNegCos}
    \!\!=\!\sum\limits_{c=1}^C\!\Big(\! \!-\frac{1}{2\tau^{c+}_i}  \text{sim}\big(\vz^c_i, [\vh^+_i]_\text{sg}\big) - \frac{1}{2\tau^{c\widetilde{+}}_i} \text{sim}\big(\vz^{c+}_i, [\vh_i]_\text{sg}\big)$}\Big. & { $\Big.\!\!\!\!\!\!\blue{+\beta\Omega(\tau^{c+}_i)\!+\!\beta\Omega(\tau^{c\widetilde{+}}_i)}\Big)$} & $\!\!\!\!\!\!$\refstepcounter{tableeqn} (\thetableeqn)\label{eq:simsiam2}$\!\!$ \\
{Barlow Twins} & { $\ell^{\ddagger}_{\text{Cross-Corr}}\! =\!\sum\limits_{c=1}^C \!\Big(\! \sum\limits_{l=1}^{d'}(1\!-\!\frac{1}{\tau^{c+}_{l}}\mathcal{C}_{ll})^2 \!\!+\! \lambda \sum\limits_{l=1}^{d'} \sum\limits_{m \neq l}^{d'} \frac{1}{\tau^{c-}_{lm}}\mathcal{C}_{lm}^2$}\Big. & {\fontsize{8}{8}\selectfont $\Big.\!\!\!\!\!\!\!\!\!\!\!\!\!\!\!\!\!\!\!\!\!\!\!\!\!\!\!\!\!\!\!\!\!\!\!\!\!\blue{+ \beta\!\sum\limits_{l=1}^{d'}\Omega(\tau^{c+}_{l})\!-\!\beta\!\sum\limits_{l=1}^{d'}\sum\limits_{m\neq l}^{d'}\Omega(\tau^{c-}_{lm})}\Big)$} & $\!\!\!\!\!\!$\refstepcounter{tableeqn} (\thetableeqn)\label{eq:bt2}$\!\!$ \\
{LGP, CAN} &  {$\ell^{\ddagger}_\text{InfoNCE}\;\;=\!\sum\limits_{c=1}^C\!\Big(\! \!-\frac{1}{\tau^{c+}_i}\text{sim}(\vz^c_i, \vz^{c+}_i)\!+\!\frac{1}{\tau^{c-}_{in^*}}\max\limits_{n=1,\cdots,N+1}\!\!\!\!\!\!\text{sim}(\vz^c_i, \vz^{c\pm}_{in})\Big.$} & {$\!\!\!\Big.\blue{+\beta\Omega(\tau^{c+}_i)\!-\!\beta\Omega(\tau^{c\pm}_{in^*})}\Big)$} & $\!\!\!\!\!\!$\refstepcounter{tableeqn} (\thetableeqn)\label{eq:infonce2}$\!\!$ \\
\bottomrule
\end{tabular}}
\vspace{-0.5cm}
\end{table*}

Our final AMCL loss is defined as $\ell^\ddagger\!=\!\sum_{c=1}^C\ell_c^\dagger$, where $\ell_c^\dagger$ represents the loss for the $c$-th head as defined in Eq.~\eqref{eq:ours}.
Table \ref{tab:loss_functions2} presents the specific implementations of our AMCL loss for various contrastive learning frameworks. The `$\ddagger$' in Eq. \eqref{eq:ntxent2}--\eqref{eq:infonce2}  indicates that these losses represent our multi-head loss versions. 

In Eq. \eqref{eq:ntxent2}, the asterisk `*' in $\tau^{c-}_{in^*}$ indicates the index $n^*\!=\!\argmax\limits_{n=1,\cdots,N}\,\text{sim}(\vz_i, \vz^-_{in})$. In Eq. \eqref{eq:simsiam2}, $\tau^{c+}_i\!=\!\sigma\big(\langle\phi(\vz^c_i), \phi([\vh^+_i]_\text{sg})\rangle\big)$ and $\tau^{c\widetilde{+}}_i\!=\!\sigma\big(\langle\phi(\vz^{c+}_i), \phi([\vh_i]_\text{sg})\rangle\big)$. In Eq. \eqref{eq:bt2}, temperatures are formed as $\tau^{c+}_l\!=\!\sigma\big(\langle\phi(\vz^c_{l:}), \phi(\vz^{c+}_{l:})\rangle\big)$ and $\tau^{c-}_{lm}\!=\!\sigma\big(\langle\phi(\vz^c_{l:}), \phi(\vz^{c+}_{m:})\rangle\big)$ where $l\neq m$ and operator `:' simply indexes and concatenates variable as $\vz_{l:}\!=\![z_{l1},\cdots,z_{lN}]^T$. In  Eq. \eqref{eq:infonce2},  $\vz^{c\pm}_{in}\!=\!\vz^{c-}_{in}$ and $\tau^{c\pm}_{in}\!=\!\tau^{c-}_{in}$ for $n=1,\cdots,N$, and  $\vz^{c\pm}_{in}\!=\!\vz^{c+}_{i}$ and $\tau^{c\pm}_{in}\!=\!\tau^{c+}_{i}$ if $n\!=\!N\!+\!1$. 
 Finally, notice that  for NT-Xent in Eq. \eqref{eq:ntxent2}, we have:
 \begin{align}
     & \qquad\!\!\!\frac{1}{\tau^{c-}_{in^*}}\!\max\limits_{n=1,\cdots,N}\!\!\!\text{sim}(\vz^c_i, \vz^{c-}_{in})\!-\!\Omega(\tau^{c-}_{in^*})\!-\!(2\pi)^{d'/2} \nonumber \\
     & \approx\log\sum_{n=1}^N\frac{1}{(2\pi)^{d'/2}(\tau^{c-}_{in})^{d'/2}}\exp\Big(\frac{1}{\tau^{c-}_{in}}\big(\text{sim}(\vz^c_i, \vz^{c-}_{in})\!-\!1\big)\Big).
      \label{eq:approx1}
 \end{align}

The same approximation (with $\vz^{c\pm}_{in}$ in place of $\vz^{c-}_{in}$) holds for InfoNCE in Eq. \eqref{eq:infonce2}. Using maximum in Eq. \eqref{eq:approx1}, Eq. \eqref{eq:ntxent2} and Eq. \eqref{eq:infonce2} are somewhat restrictive as the soft-maximum, depending on the temperature, will return the maximum similarity or interpolation over top few similarities close to maximum. Thus, soft-maximum will tackle a group of negative samples closest to the anchor. Indeed, we could use the above soft-maximum in place of maximum but then we have no easy way of recovering the temperature $\tau^{c-}_{in^*}$. Thus, in our experiments we observed that the best choice is to apply $\sum_{k=1}^\kappa \frac{1}{\kappa\tau^{c-}_{in^*_{k}}}\big[\topmaxkappa\limits_{n=1,\cdots,N }\, \text{sim}(\vz^c_i, \vz^{c-}_{in})\big]_{k}$ which is the average over top-$\kappa$ largest similarities. As the top-$\kappa$ maximum operation returns also corresponding indexes  $n^*_{1},\cdots,n^*_{\kappa}$, the temperature regularization can  be easily computed as $\Omega\big(\{n^*_{k}\}_{k=1}^\kappa\big)=\log(\tau^{c-}_{in^*_{1}}\cdot\ldots\cdot\tau^{c-}_{in^*_{\kappa}})+\sum_{k=1}^\kappa\frac{1}{\tau^{c-}_{in^*_{k}}}$. 
The derivation of our loss function is presented in Appendix~\ref{sec:deriv}.

\subsection{Discussion}
\noindent\textbf{Multiple projection heads \vs a wider MLP projection head.} 
Multiple projection heads offer a significant advantage by creating different sets of features for each pair of images~\cite{tao2019deep}. This helps build a varied representation that captures various aspects of the input, making the model more resilient and adaptable to different transformations. Furthermore, each projection head's ability to specialize in learning specific patterns or features becomes especially useful when dealing with diverse augmentations that require focused attention on distinct content within the images.
In the context of contrastive learning, where diverse augmentation strategies and varying intra-sample similarity introduce more variability in positive pairs, using multiple projection heads excels. This approach effectively manages the resulting variability by enabling the learning of diverse representations. As a result, the model becomes skilled at navigating the intricacies of the data, demonstrating its adaptability to the complex nature of the input.
On the other hand, the wider MLP projection head takes a distinct approach by consolidating information from various aspects of the input into a single representation. This proves advantageous when a thorough understanding of the data is crucial, and the relationships between features play a vital role~\cite{chen2020simple}.

\begin{figure}[tbp]
    \centering 
\begin{subfigure}{0.245\linewidth}
  \includegraphics[width=\linewidth]{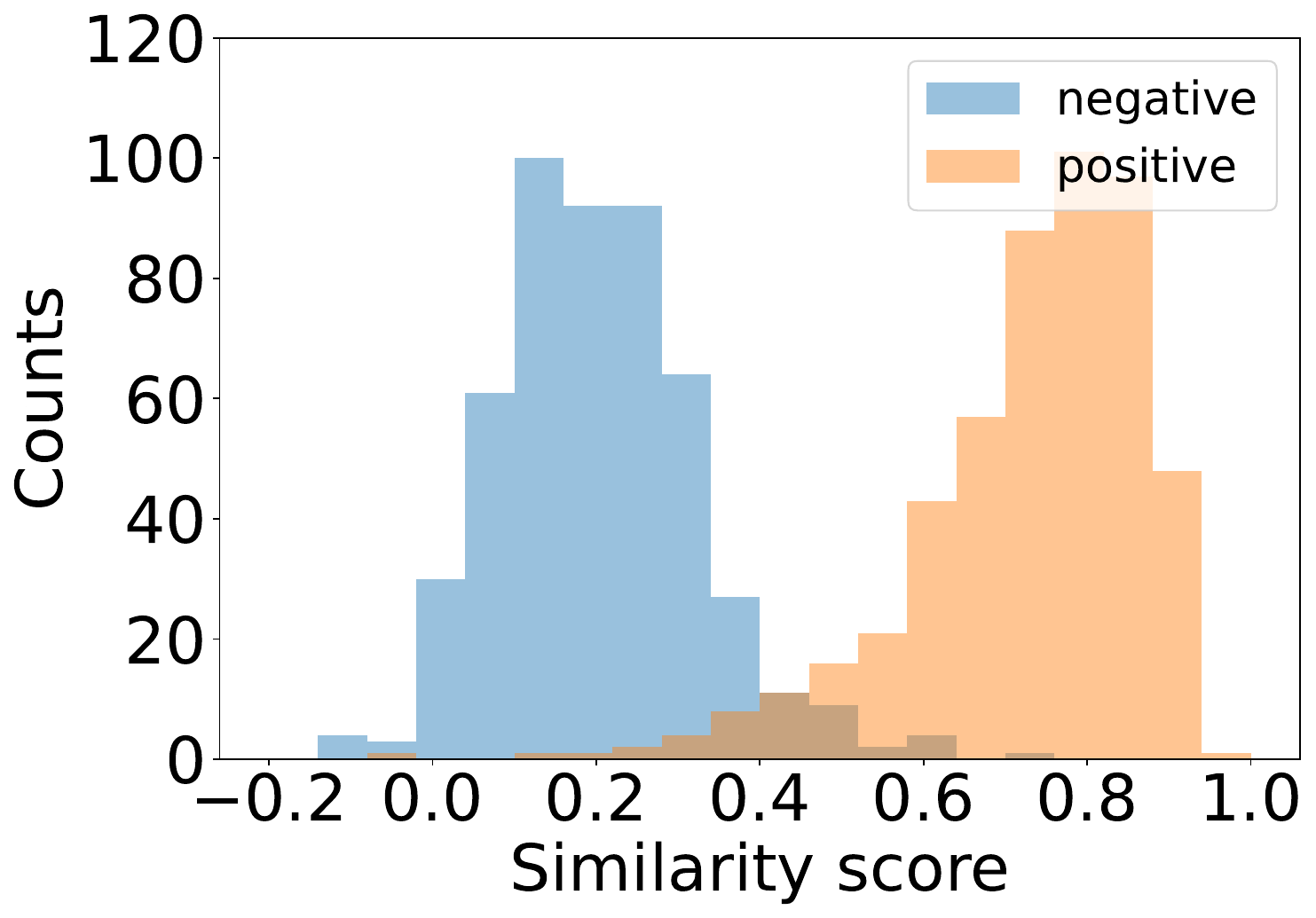}
  \caption{head sim. (baseline)}
  \label{fig:1}
\end{subfigure}\hfil 
\begin{subfigure}{0.245\linewidth}
  \includegraphics[width=\linewidth]{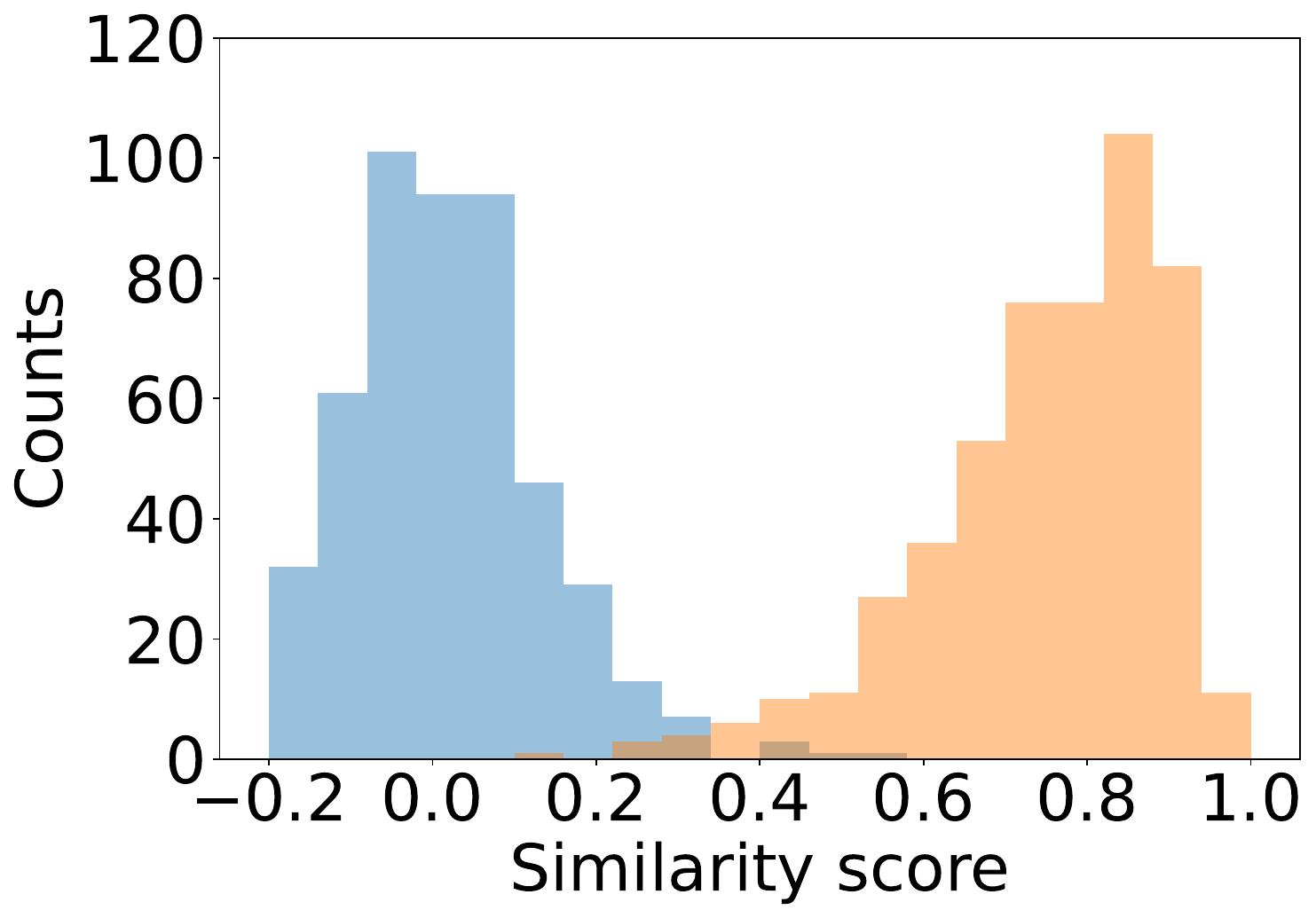}
  \caption{head sim. (ours)}
  \label{fig:3}
\end{subfigure}\hfil
\begin{subfigure}{0.245\linewidth}
  \includegraphics[width=\linewidth]{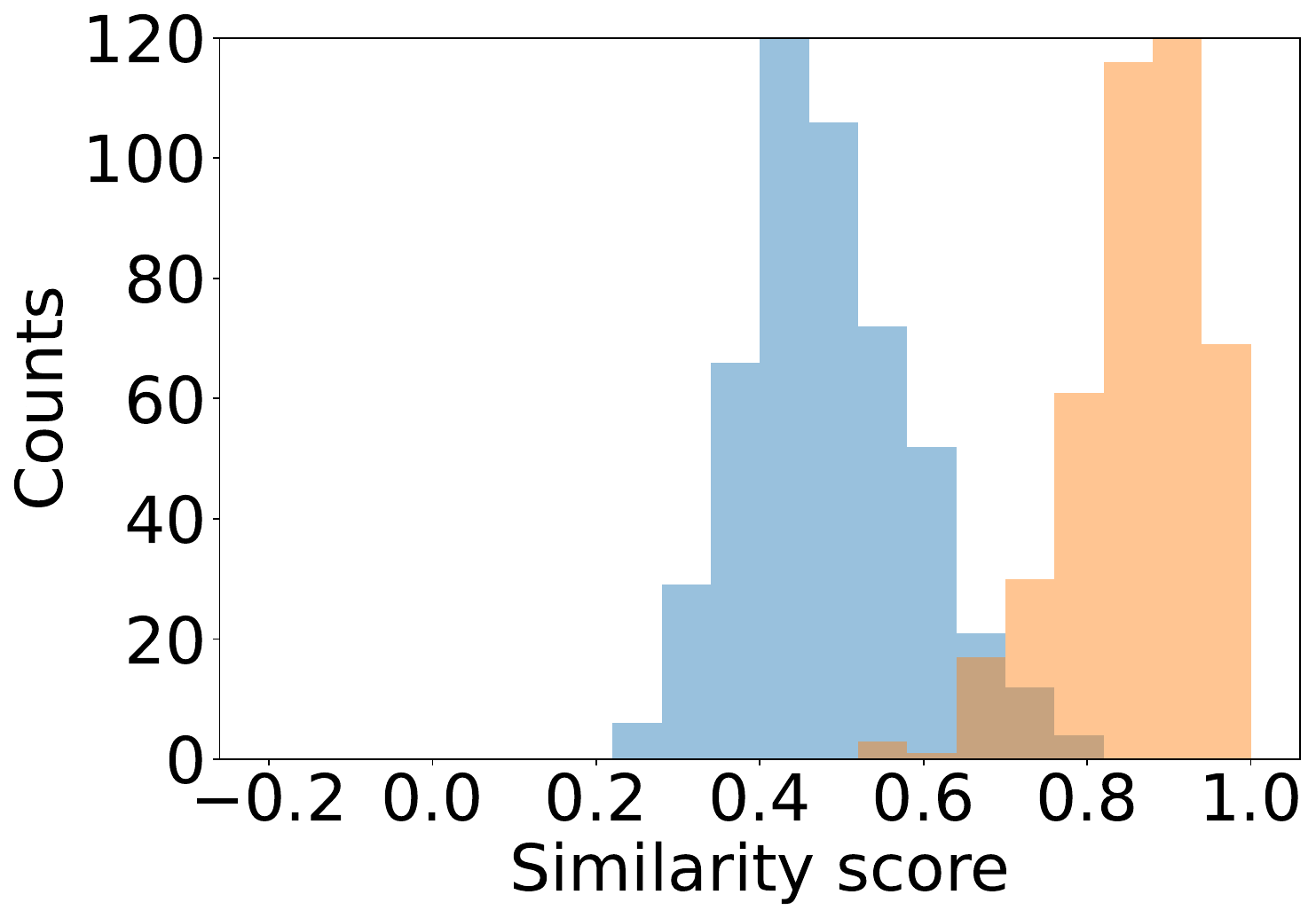}
  \caption{feat. sim. (baseline)}
  \label{fig:7}
\end{subfigure}\hfil 
\begin{subfigure}{0.245\linewidth}
  \includegraphics[width=\linewidth]{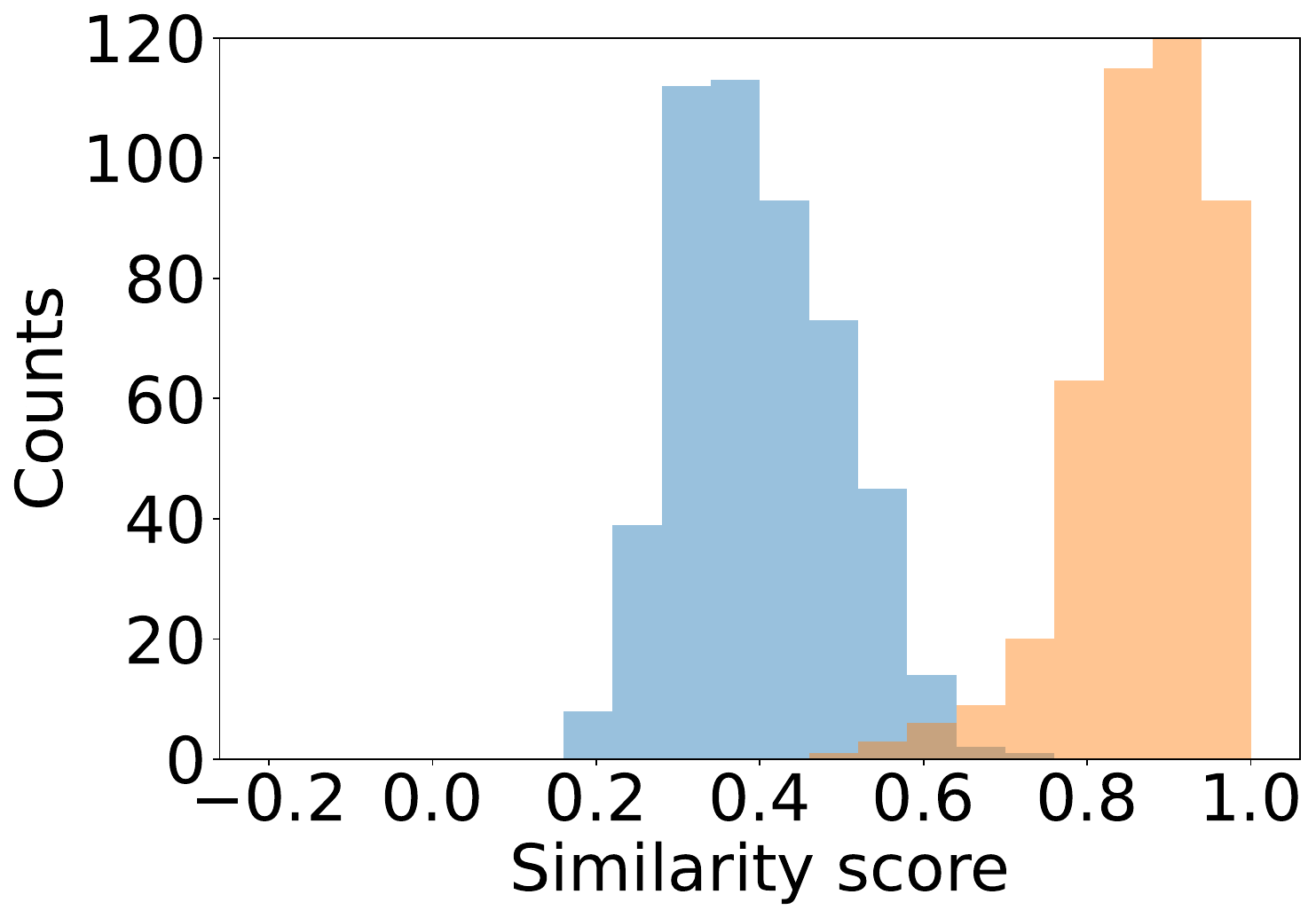}
  \caption{feat. sim. (ours)}
  \label{fig:8}
\end{subfigure}
\vspace{-0.2cm}
\caption{Distribution of similarity scores 
for positive and negative pairs. The baseline uses one projection head and constant temperature, while our method has multiple projection heads and adaptive temperature. We use SimCLR for pre-training with ResNet-18 on STL-10. After pre-training, we choose 500 positive pairs and 500 negative pairs from the validation to compute the cosine similarity. In (a) and (b), similarity score (temperature scaled) is computed between the 128-dim features extracted from the projection head(s). In (c) and (d), cosine similarity score is computed between the 512-dim features extracted from the backbone after removing the project heads.} 
\label{fig:images}
\vspace{-0.5cm}
\end{figure}

\noindent\textbf{Visualization of pair similarity distributions.} In Fig. \ref{fig:images}, we draw the similarity distributions of negative pairs and positive pairs, under the baseline (1 head + constant temperature) and our method (multiple heads + adaptive temperature). When we use the average similarity across the output from the multiple heads, shown in Fig. \ref{fig:images}(a) and (b), we can clearly observe better separability brought by our method. It indicates that our method allows for more effective similarity learning of the positive and negative pairs. On the other hand, if we compute the cosine similarity between features extracted right after the backbone, shown in Fig. \ref{fig:images}(c) and (d), better separability can again be observed. It illustrates that better similarity learning further benefits representation learning, finally leading to improved linear probing performance. 

\noindent\textbf{Connecting our loss function to existing contrastive learning methods.} 
Our loss function consists of three terms: (i) positive temperature-weighted similarities for positive pairs (ii) negative temperature-weighted similarities for negative pairs, and (iii) a regularization term for positive and negative temperatures. As identified in previous works~\cite{wang2020understanding, wang2021understanding}, the alignment (closeness) of features from positive pairs and the uniformity of the induced distribution of the (normalized) features on the hypersphere are the two key properties in contrastive loss. 
Our loss function also optimizes these properties and further improves the contrastive learning performance through parameterized pair-wise temperature via re-weighting the positive and negative similarities. Our loss function is a more general form, and when we set the temperature to be a global constant, the constant regularization term no longer affects optimization, and thus the loss function reduces to the traditional contrastive loss.

\noindent\textbf{Physical meaning of the regularization term $\Omega(\cdot)$ in Eq.~\eqref{eq:ours}}. 
The regularization term consists of regularizing both the positive and negative temperatures. During optimisation, the $\log\tau$ term encourages lower positive temperatures and higher negative temperatures, whereas the reverse function term $\frac{1}{\tau}$ is in favour of higher positive temperatures and lower negative temperatures. Hence this regularization term balances the learning of positive and negative temperatures. 
As we jointly optimize network parameters and the temperature via MLE, this problem naturally becomes decomposed in the maximization of similarities for positive pairs (minimization for negative pairs) weighted by the temperature. However, if $\tau$ was to reach 0 for positive pairs, one would attain a trivial solution. $\Omega(\cdot)$ prevents that trivial solution. Intuitively, one can be very certain in similarity of sample pair but there is a price to pay for that certainty, imposed by $\Omega(\cdot)$ resulting from the Welsch  function. This  $\Omega(\cdot)$ expresses the prior belief or preference on temperatures. We provide its mathematical insights including connecting temperature to uncertainty in Appendix~\ref{sec:deriv}. More discussions can be found in Appendix~\ref{app:discussion}.

\section{Experiments}

We choose popular datasets that are widely used in evaluating the SSL models, including CIFAR-10, CIFAR-100, STL-10, Tiny-ImageNet and ImageNet. 
The dataset details are provided in Appendix \ref{app:data}. 
Below, we describe the experimental setup and evaluations.

\subsection{Setup}

We conduct experiments on the aforementioned datasets following practices outlined by~\cite{huang2022towards}. We consider five different types of transformations for data augmentations: random cropping, random Gaussian blur, color dropping (\ie, randomly converting images to grayscale), color distortion, and random horizontal flipping. 
For pre-training, we employ Resnet-18 (R18)~\cite{he2016deep} for SimCLR~\cite{chen2020simple}, MoCo~\cite{he2020momentum}, SimSiam~\cite{chen2021exploring}, and Barlow Twins~\cite{zbontar2021barlow} on CIFAR-10, CIFAR-100 and STL-10. For Tiny-ImageNet and ImageNet, we use 
Resnet-50 (R50) variants, ViT-B and ViT-L. Other settings, such as the architecture of the projection head, remain consistent with the original algorithm configurations. 
We select $C$ projection heads in range from 2 to 6. We set the temperature bounds
($\eta$ and $\iota$ of sigmoid) in range  [$1e-5$, 2] on smaller datasets and  [$1e-5$, 5] for ImageNet and  Tiny-ImageNet. 
For Barlow Twins, the regularization parameter $\lambda=5e\!-\!3$. The temperature regularization parameter $\beta$ is varied from $1e-5$ to 10. 
Each model is trained with a batch size of 512 and 1000 epochs for small datasets, \eg, STL-10; for large-scale datasets, \eg, Tiny-ImageNet and ImageNet, we train for up to 800 epochs. 
We also evaluate ViT-B and ViT-L pre-trained for 1600 epochs on ImageNet and assess their generalizability, \eg, on object detection and segmentation.
To assess the quality of the encoder, we follow the KNN evaluation protocol~\cite{wu2018unsupervised} on small datasets. For large datasets, we use linear probing. For MIM-based methods such as CAN\footnote{https://github.com/bwconrad/can} and LGP\footnote{https://github.com/VITA-Group/layerGraftedPretraining\_ICLR23}, we select the standard ViT-B and ViT-L as the backbone encoder, with a token size $16\!\times\!16$. Other settings, including projection head architecture and hyperparameters, followed the original algorithm configurations. Both models are evaluated using a linear probe.

\subsection{Evaluation}

\begin{table*}[tb]
\begin{center}
\caption{Impact of AMCL when applied to popular and state-of-the-art SSL methods. On CIFAR-10, CIFAR-100, Tiny-ImageNet, and ImageNet, models are first pretrained for 1,000, 1,000, 800, and 100 epochs, respectively, and then evaluated using linear probing. \bkbone{Backbones} are highlighted. All reported results are averages obtained from 5 runs on each dataset for each method (including baselines).
}
\vspace{-0.5cm}
\resizebox{\linewidth}{!}{
\begin{tabular}{ l c cc  cc  cc  cc cc cc c}
\toprule
\multirow{2}{*}{Datasets} & & \multicolumn{12}{c}{Accuracy} & Avg.\\
\cline{3-14}
 & & \multicolumn{2}{c}{SimCLR} & \multicolumn{2}{c}{MoCo} & \multicolumn{2}{c}{SimSiam}  & \multicolumn{2}{c}{B.Twins} & \multicolumn{2}{c}{CAN}&  \multicolumn{2}{c}{LGP} & gain\\
\midrule
\multirow{2}{*}{CIFAR-10} 
& Baseline & 89.9  &\multirow{2}{*}{\bkbone{R18}}&  90.4  &\multirow{2}{*}{\bkbone{R18}}& 90.7  &\multirow{2}{*}{\bkbone{R18}}& 87.4 &\multirow{2}{*}{\bkbone{R18}}& - &&-  && \multirow{2}{*}{\textcolor{red}{\bf $\uparrow$2.43}} \\
& \red{\textbf{+AMCL}}  & \textbf{92.2} & & \textbf{92.9} & & \textbf{93.0}  && \textbf{90.0} && - &&-  && \\
\hline
\multirow{2}{*}{CIFAR-100} 
& Baseline & 57.6    &\multirow{2}{*}{\bkbone{R18}}& 64.4  &\multirow{2}{*}{\bkbone{R18}}& 63.6 &\multirow{2}{*}{\bkbone{R18}}& 58.2 &\multirow{2}{*}{\bkbone{R18}}& - && -&& \multirow{2}{*}{\textcolor{red}{\bf $\uparrow$4.58}} \\
& \red{\textbf{+AMCL}} & \textbf{61.8} & &\textbf{69.3} && \textbf{68.9} && \textbf{62.1} && - && -&& \\
\hline
\multirow{2}{*}{Tiny-ImageNet} 
& Baseline & 48.1  &\multirow{2}{*}{\bkbone{R50}}& 46.4  &\multirow{2}{*}{\bkbone{R50}}& 46.7  &\multirow{2}{*}{\bkbone{R50}}& 46.8 &\multirow{2}{*}{\bkbone{R50}}& 53.2 &\multirow{2}{*}{\bkbone{ViT-B}}& 56.7 &\multirow{2}{*}{\bkbone{ViT-B}}& \multirow{2}{*}{\textcolor{red}{\bf $\uparrow$1.65}} \\
& \red{\textbf{+AMCL}} & \textbf{50.0} && \textbf{47.8}  & & \textbf{49.0}  && \textbf{48.3} && \textbf{54.9} && \textbf{57.8}&& \\
\hline
\multirow{2}{*}{ImageNet} 
& Baseline & 66.5 &\multirow{2}{*}{\bkbone{R50}} & 67.4 & \multirow{2}{*}{\bkbone{R50}} & 68.1 &\multirow{2}{*}{\bkbone{R50}}& 70.0& \multirow{2}{*}{\bkbone{R50}} & 70.5 &\multirow{2}{*}{\bkbone{ViT-B}}& 73.8&\multirow{2}{*}{\bkbone{ViT-B}}& \multirow{2}{*}{\textcolor{red}{\bf $\uparrow$1.67}} \\
& \red{\textbf{+AMCL}} & \textbf{68.1} & & \textbf{69.3}  && \textbf{69.6}  && \textbf{72.7} && \textbf{71.6} & & \textbf{75.0}& & \\
\bottomrule
\end{tabular}}
\label{full-table}
\end{center}
\vspace{-0.5cm}
\end{table*}

\begin{table}[tb]
    \centering
    \caption{A comprehensive/fair comparison on ImageNet \vs SOTA. \bkbone{Backbones} are highlighted. All models are pre-trained for 1600 epochs. We report top-1 performance after end-to-end fine-tuning.}
    \vspace{-0.3cm}
    \begin{tabular}{lccccc}
        \toprule
         & \multirow{2}{*}{MAE} & \multirow{2}{*}{MoCo} &  MoCo & \multirow{2}{*}{LGP} & LGP\\
         & & & \red{\textbf{+AMCL}} & & \red{\textbf{+AMCL}}\\
        \midrule
        \bkbone{ViT-B}/16& 83.6& 83.2& \textbf{84.1} ({\textcolor{red}{\bf $\uparrow$0.9}}) &83.9 & \textbf{84.7} ({\textcolor{red}{\bf $\uparrow$0.8}})\\
        \bkbone{ViT-L}/16& 85.9& 84.1& \textbf{85.3} ({\textcolor{red}{\bf $\uparrow$1.2}}) &85.9 & \textbf{87.4} ({\textcolor{red}{\bf $\uparrow$1.5}})\\
        \bottomrule
    \end{tabular}
    \label{tab:imgnet}
    \vspace{-0.2cm}
\end{table}

{\bf AMCL consistently improves popular and state-of-the-art SSL methods.} We apply AMCL to SimCLR, MoCo, SimSiam, Barlow Twins, CAN, and LGP. As shown in Table \ref{full-table}, on CIFAR-10, CIFAR-100, Tiny ImageNet and ImageNet datasets, average improvements across the baselines are 2.43\%, 4.58\%, 1.65\%, 1.67\%, respectively. 
In the case of MIM-based methods such as CAN and LGP, our multi-head approach improves them by $1.7\%$ and $1.1\%$, respectively, on Tiny-ImageNet. The improvements over CAN and LGP on ImageNet are $1.1\%$ and $1.2\%$, respectively.
We provide more results on ImageNet to compare with SOTA in Table~\ref{tab:imgnet}. All models are pre-trained for 1600 epochs, and are evaluated by end-to-end fine-tuning. We report top-1 performance after fine-tuning. AMCL boosts popular SSL methods by 1\% on average.

\begin{figure}[tb]
    \begin{minipage}[t]{0.48\linewidth}
\centering
\begin{subfigure}{0.50\linewidth}
  \includegraphics[width=\linewidth]{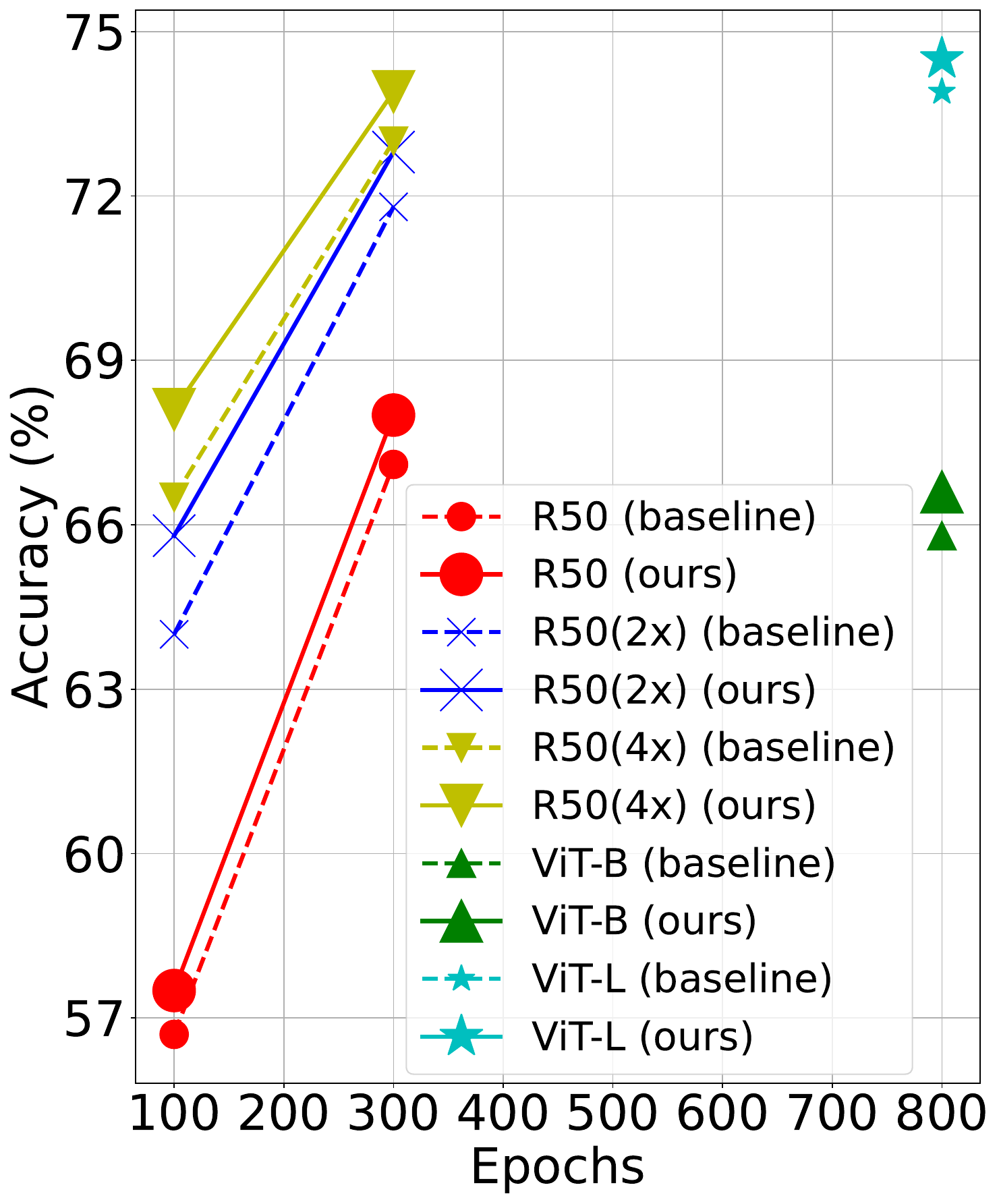}
\end{subfigure}\hfil 
\begin{subfigure}{0.50\linewidth}
  \includegraphics[width=\linewidth]{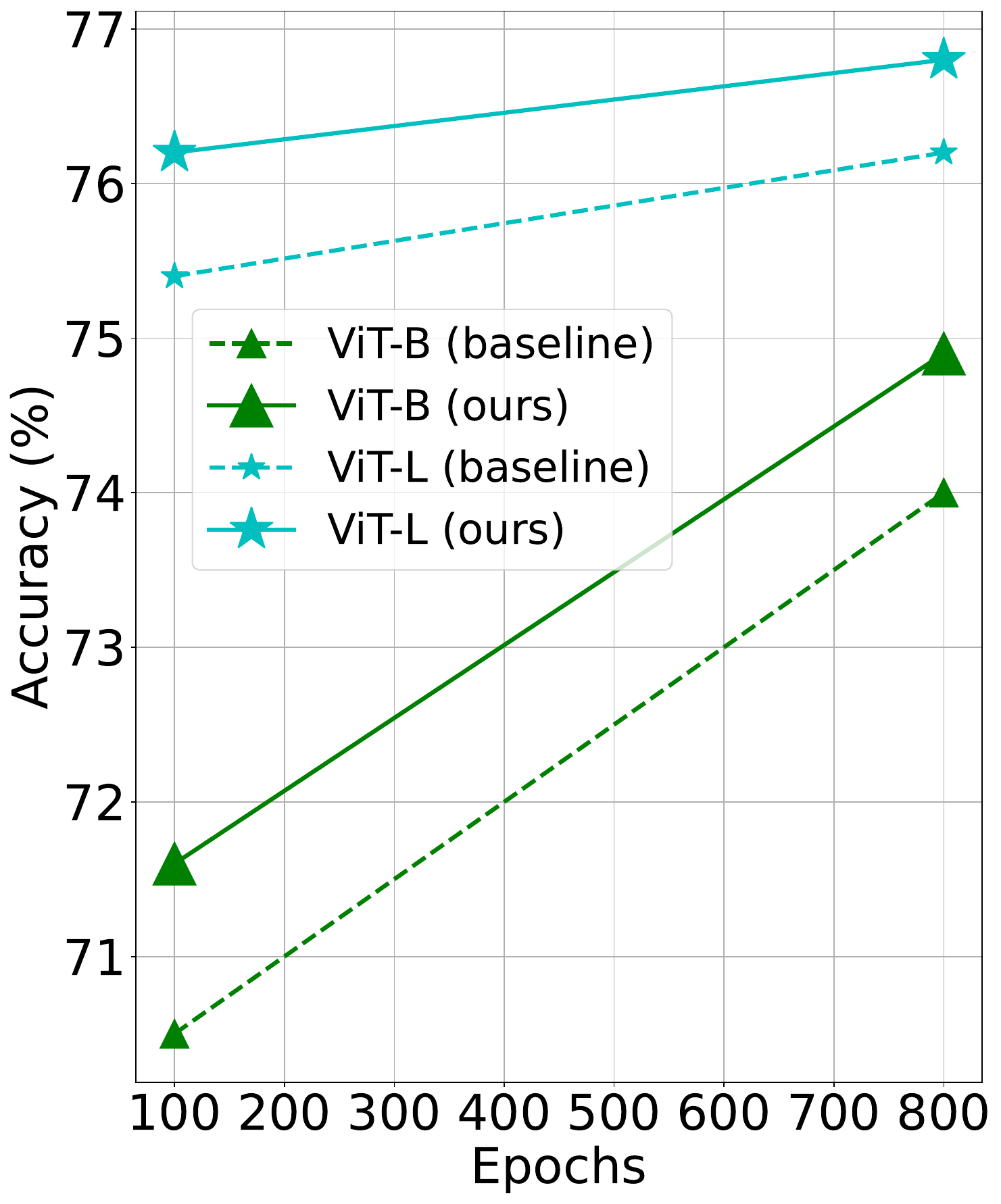}
\end{subfigure}
  \label{fig:backbone-simclr-can}
\caption{Impact of different backbones and training epochs on AMCL for (\textbf{left}:) SimCLR and (\textbf{right}:) CAN on the ImageNet dataset. All the reported accuracies use a linear probe.}
\label{Fig:encoder}
\end{minipage}\hspace{0.2cm}
\raisebox{\dimexpr\height-\baselineskip}{%
\begin{minipage}[t]{0.48\linewidth}
\centering
\begin{subfigure}{0.5\linewidth}
  \includegraphics[width=0.88\textwidth]{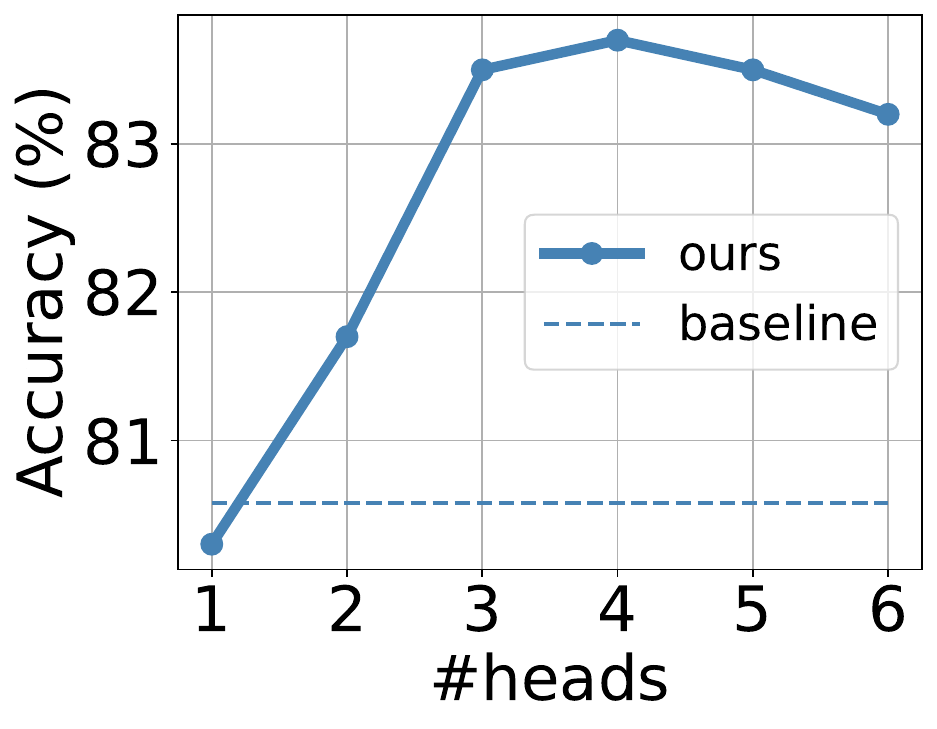}
\end{subfigure}\hfil 
\begin{subfigure}{0.5\linewidth}
  \includegraphics[width=\textwidth]{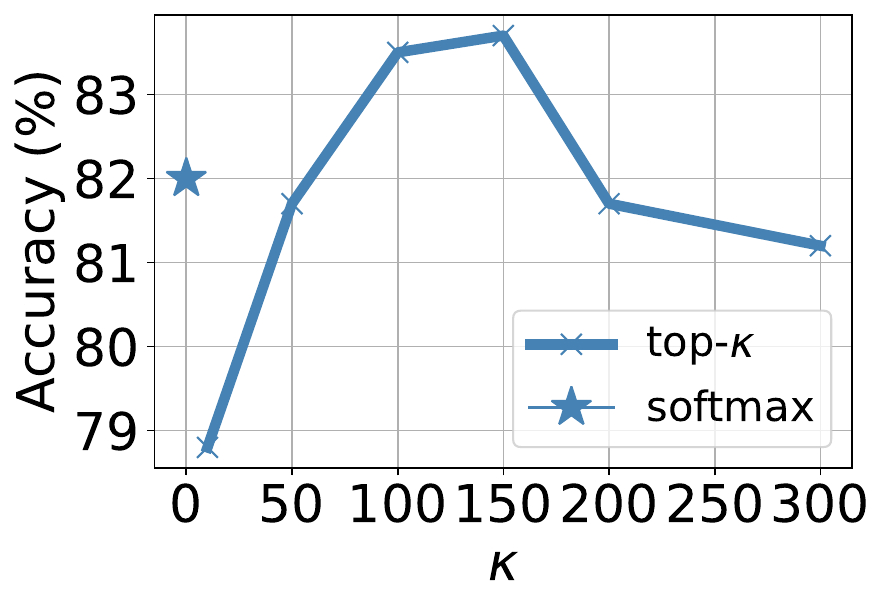}
\end{subfigure}
\caption{Hyperparameter sensitivity analysis. 
(\textbf{left}:) number of projection heads. (\textbf{right}:) Evaluation of top-$\kappa$ similarities 
among negative pairs on STL-10. 
``Softmax'' involves including all negative pair similarities (Eq.~\eqref{eq:jump} in Appendix~\ref{sec:deriv}). We use Resnet-18 as backbone with SimCLR. The dashed line is the baseline result with 1 projection head, constant temperature.}
\label{fig:heads-topk}
\end{minipage}}
\end{figure}

\begin{table}[tb]
	\begin{minipage}{0.44\linewidth}
		\caption{Comparing adaptive temperature with two SOTA temperature methods on STL-10.
  Baseline uses one projection head and constant temperature, and the rest methods use 3 heads. 
  {TaU~\cite{zhang2021temperature} 
  views temperature as uncertainty, and TS~\cite{kukleva2023temperature} uses cosine schedule for temperature.}
  }
		\label{stl10-attn}
		\centering
		\begin{tabular}{l c l c c}
\toprule
& Baseline & \red{\textbf{+AMCL}} & TaU & TS\\
\midrule
10& 61.0 & \textbf{71.1} \textcolor{red}{\bf $\uparrow$10.1}& 64.3& 64.3\\
20 & 68.5 & \textbf{76.4} \textcolor{red}{\bf $\uparrow$7.9}& 70.4& 71.3\\
50 & 73.5& \textbf{79.4} \textcolor{red}{\bf $\uparrow$5.9}& 74.6& 76.4\\
100& 76.5& \textbf{81.4} \textcolor{red}{\bf $\uparrow$4.9}& 77.7& 78.4\\
200& 78.6& \textbf{81.9} \textcolor{red}{\bf $\uparrow$3.3}& 79.4& 80.0\\
500& 80.6& \textbf{83.7} \textcolor{red}{\bf $\uparrow$3.1}& 80.0& 81.7\\
\bottomrule
\end{tabular}
	\end{minipage}\hfill
	\begin{minipage}{0.5\linewidth}
		\centering
\includegraphics[width=0.9\linewidth]{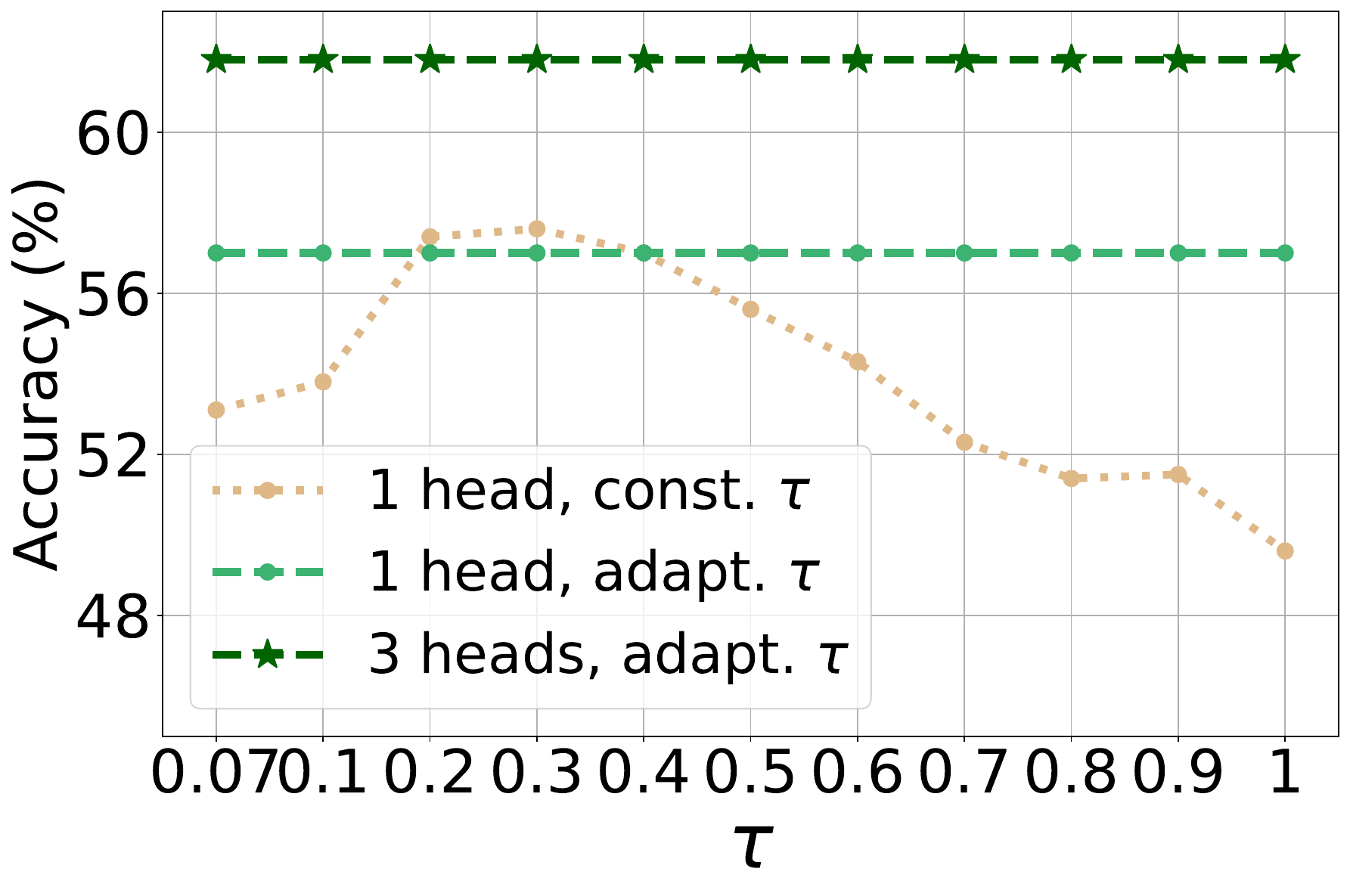}
  \captionof{figure}{Comparing adaptive temperature in multi-head ($C\!=\!3$) versus single-head ($C\!=\!1$) scenarios on CIFAR-100 using SimCLR with a ResNet-18 backbone. The single-head model with a constant temperature performs best at 0.2 or 0.3. For a single head, constant and adaptive temperatures show comparable performance.
  }
  \label{fig:temp-eval}
	\end{minipage}
\vspace{-0.5cm}
\end{table}

\noindent{\bf Effectiveness of AMCL under different backbones and training epochs.} 
We evaluate model capacity using the ImageNet dataset. We choose R50 with three different hidden layer widths (width multipliers of $1\times$, $2\times$, and $4\times$), ViT-B, and ViT-L, which are widely used in SSL.
In Fig. \ref{Fig:encoder}, under each epoch number, our AMCL yields consistent improvement to various backbones for SimCLR and CAN.  
Additionally, we observe that model capacity heavily depends on the choice of backbone. 
For ViT-L backbone encoder, SimCLR achieves the highest linear probe performance on ImageNet at $73.9\%$. When combined with AMCL, accuracy of SimCLR {further} increases by  $0.6\%$.

\noindent{\bf Impact of the number of projection heads} is presented in  Fig.~\ref{fig:heads-topk} (left).  
R18 is used as backbone, coupled with SimCLR. Adaptive temperature is always used. 
On STL-10, 3-5 projection heads are most effective, while other numbers also beat the baseline (except for $C\!=\!1$). Considering the performance gain and the computational cost, we use 3 heads in our paper. 
We notice a 1\% performance drop when the number of heads exceeds 4, likely due to slight overfitting.
Note that this is not a hyperparameter selection process but rather hyperparameter sensitivity evaluation.
Instead, we choose hyperparameters using Hyperopt~\cite{bergstra2015hyperopt} on the validation set of each dataset.

\noindent{\bf TopMax-$\kappa$ \vs Softmax in loss function.}  In Fig.~\ref{fig:heads-topk} (right), 
we compare using Top-$\kappa$ {and} Softmax when selecting negative pairs. 
Using Top-$\kappa$ is slightly better than Softmax when $\kappa=100, 150$. In practice, $\kappa$ is chosen by Hyperopt~\cite{bergstra2015hyperopt} on the validation set of each dataset.

\begin{figure}[tb]
\centering
  \centering
\begin{subfigure}{0.48\linewidth}
  \includegraphics[width=0.98\textwidth]{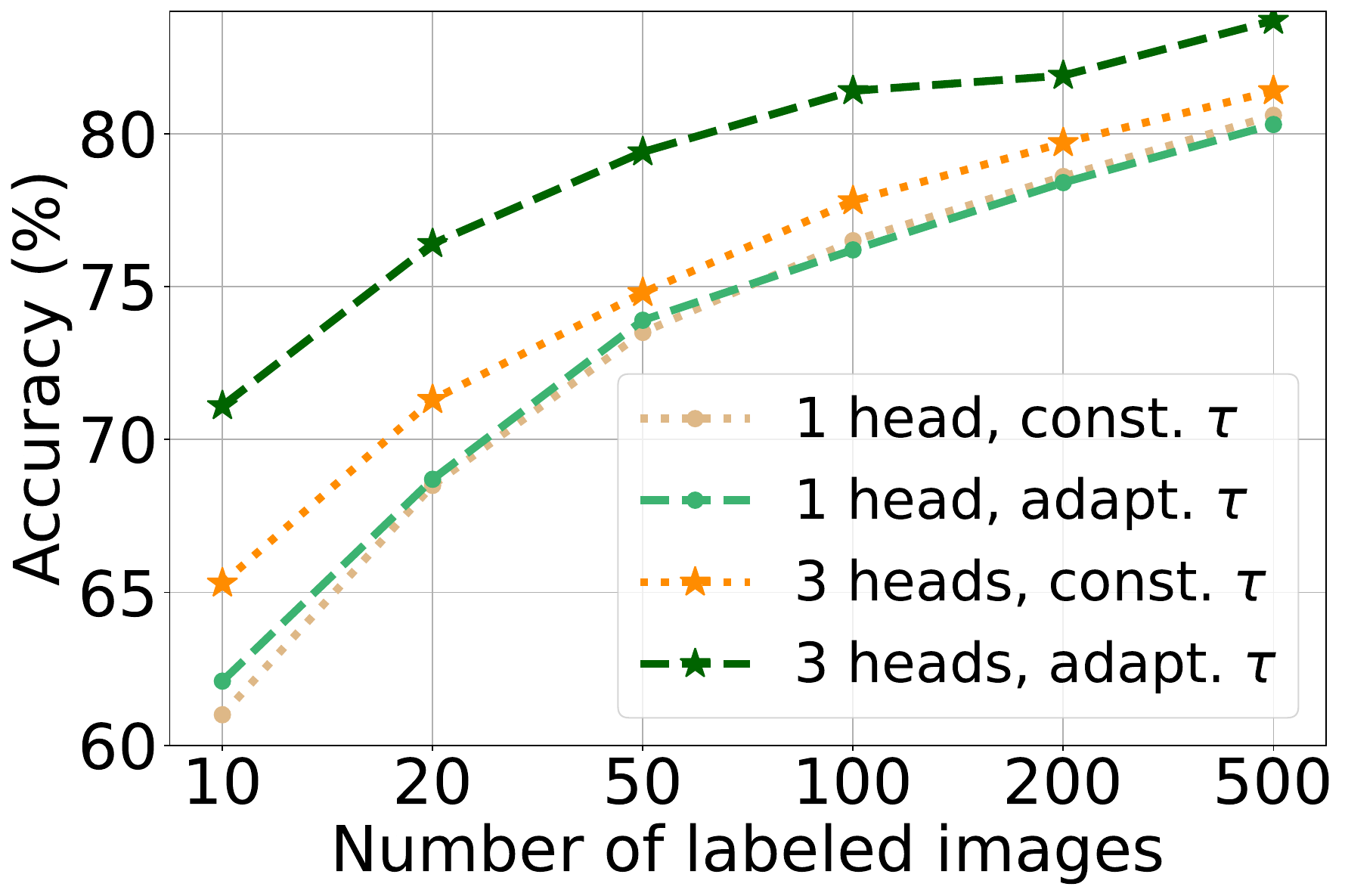}
\end{subfigure}\hfill
\begin{subfigure}{0.48\linewidth}
  \includegraphics[width=\textwidth]{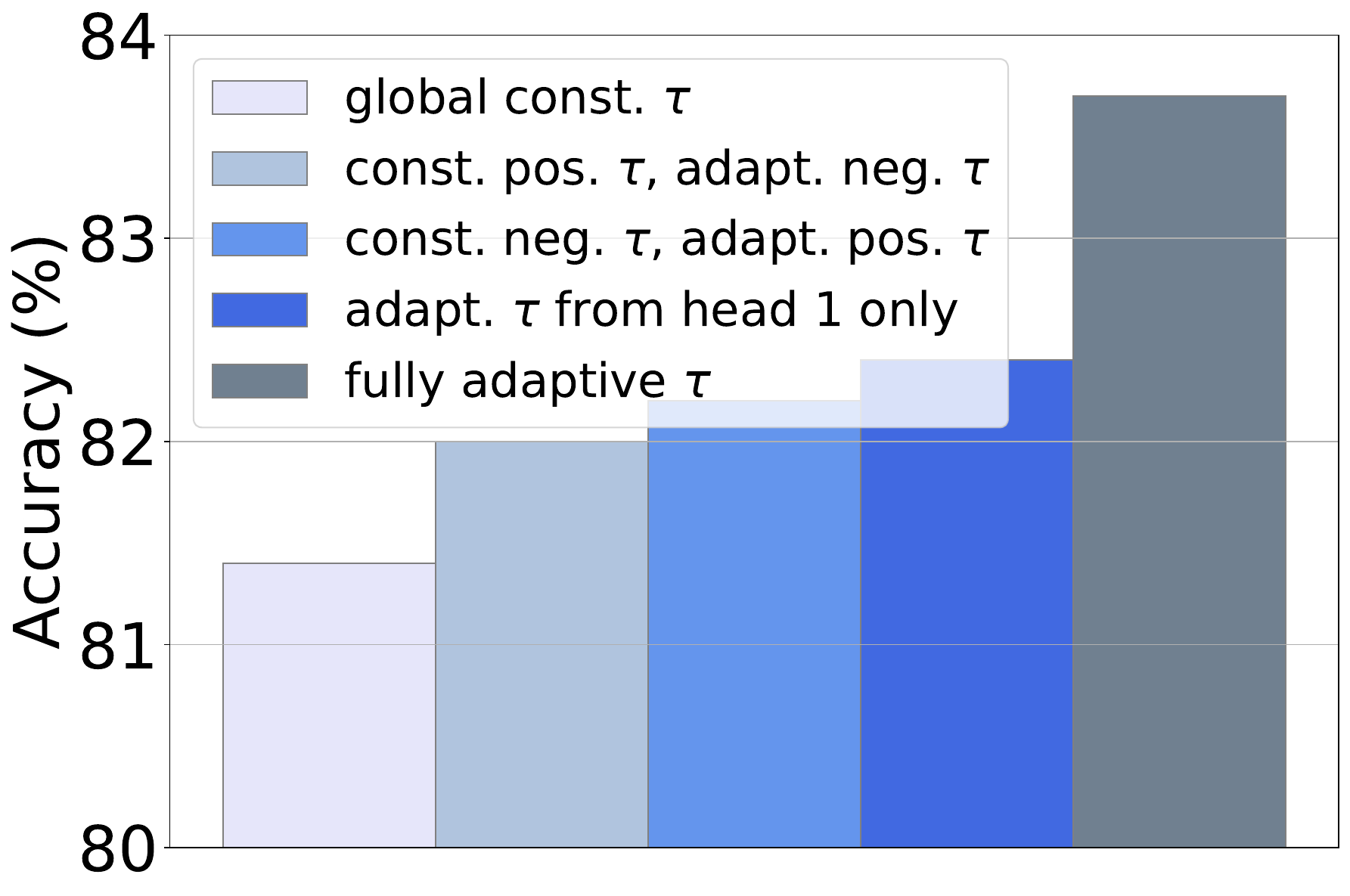}
\end{subfigure}
\caption{Comparing variants of adaptive temperature on STL-10. We choose SimCLR. The pre-trained model is linear-probed with various numbers of labeled data. (\textbf{left}:) multi-head ($C\!=\!3$) \vs single-head ($C\!=\!1$).
(\textbf{right}:) constant temperature \vs adaptive temperature. The Number of heads is 3. ``Const. pos. $\tau$, adapt. neg. $\tau$'' uses constant temperature for positive pairs and adaptive for negative ones. Analogy goes for ``Const. neg. $\tau$, adapt. pos. $\tau$''. ``Adapt. $\tau$ from head 1 only'' copies adaptive temperatures from the first head to the other heads.}
\label{fig:temp-stl10}
\vspace{-0.5cm}
\end{figure}

\noindent{\bf Impact of number of labeled data for linear probing.}  
We use SimCLR for pre-training and linear-probe a standard R18 model (with random initialization) on the training set of STL-10.
We then train a logistic regression model with various numbers of labeled images: 10, 20, 50, 100, 200, and all 500 examples per class. Results are presented in Table \ref{stl10-attn} and Fig.~\ref{fig:temp-stl10} (left). We have two observations. \textit{First}, AMCL consistently improves SimCLR under different numbers of labeled data. Under 10, 50, and 200 labeled images, the improvement is 10.1\%, 5.9\%, and 3.3\%, respectively. \textit{Second}, compared with the fully supervised baseline with 73.3\% accuracy, linear probing with 20 samples per class (1/25 of the whole labeled data) with our method is already superior (76.4\%). This clearly demonstrates the advantages of self-supervised pretraining and our method.

\noindent{\bf Comparing with temperature variants and state-of-the-art temperature schemes.} 
In Fig.~\ref{fig:temp-stl10} (right), we compare the proposed adaptive temperature (fully adaptive $\tau$) with four variants, including making temperature of negative/positive pairs constant, copying adaptive temperature from one head to the others, and having global temperature. It is clear from the figure that our method is the best. In Table~\ref{stl10-attn}, we compare adaptive temperature with temperature as uncertainty (TaU) \cite{zhang2021temperature} and temperature cosine schedule (TS) \cite{kukleva2023temperature} under multiple heads, where adaptive temperature is also superior. In fact, the cosine temperature scheme is not adaptive to individual pairs, while `temperature as uncertainty' is directly dependent on features not similarity. Being adaptive to individual pairs, their similarity and projection heads, our adaptive temperature is very well optimized under the derived loss function and thus exhibits very competitive performance (see also Fig.~\ref{fig:temp-eval}).

\noindent\textbf{Multi-head outperforms a wider MLP projection head.} 
To evaluate the efficacy of employing multiple projection heads versus a wider MLP projection head, we conducted two sets of experiments: (i) with a global constant temperature and (ii) with our adaptive temperature on STL-10, employing both SimCLR and Barlow Twins. We systematically increased the number of learnable weights for the wider projection head until it surpassed three times the number of weights in our multi-head settings. We then selected {\it the best-performing model} for comparison. Our results demonstrate that, under both constant and adaptive temperature conditions, our multi-head approach (with $C\!=\!3$, 128-dim for SimCLR, and 2048-dim for Barlow Twins) outperforms {\it the best wider MLP projection head} (512-dim for SimCLR and 8192-dim for Barlow Twins) by approximately 2.5\% and 1.6\% for Barlow Twins and SimCLR, respectively.

\begin{table}[tb]
\begin{center}
\caption{Comparing AMCL with baselines under various numbers of augmentations. 
%
%
We report linear probing accuracy on CIFAR-100, where (a), (b), (c), (d), and (e) correspond to random cropping, Gaussian blur, color dropping (\eg, converting images to grayscale), color distortion, and random horizontal flipping, respectively. $\checkmark$ denotes the corresponding augmentation is applied. Average improvement is shown in red.}\label{cifar100-aug}
\vspace{-0.3cm}
\begin{tabular}{ c c c  c  c  l c c c  c  c}
\toprule
\multicolumn{5}{c}{Augmentations} & & \multicolumn{4}{c}{Accuracy} & Avg.\\
\cline{1-5}
\cline{7-10}
 (a) & (b) & (c) & (d) & (e) & & SimCLR & MoCo & SimSiam & B.Twins  & gain\\
\midrule
\multirow{2}{*}{$\checkmark$} & & & & & Baseline & 26.3 & 39.9 & 26.4 & 33.7 & \multirow{2}{*}{\textcolor{red}{\bf $\uparrow$0.69}}\\
& & & & & \red{\textbf{+AMCL}} & \textbf{27.0} & \textbf{40.3} & \textbf{27.0} & \textbf{34.7} & \\
\hline
\multirow{2}{*}{$\checkmark$} & \multirow{2}{*}{$\checkmark$}  & & & & Baseline & 28.0 & 40.2 & 26.6  & 34.2 & \multirow{2}{*}{\textcolor{red}{\bf $\uparrow$0.80}}\\
& & & & & \red{\textbf{+AMCL}} & \textbf{29.0} & \textbf{40.9} & \textbf{27.3} & \textbf{35.1} &\\
\hline
\multirow{2}{*}{$\checkmark$} & \multirow{2}{*}{$\checkmark$} & \multirow{2}{*}{$\checkmark$}  & & & Baseline & 44.4 & 57.3 & 51.5 & 49.8 & \multirow{2}{*}{\textcolor{red}{\bf $\uparrow$2.74}}\\
& & & & & \red{\textbf{+AMCL}} & \textbf{46.2} & \textbf{61.0} & \textbf{55.4} & \textbf{51.3} & \\
\hline
\multirow{2}{*}{$\checkmark$} & \multirow{2}{*}{$\checkmark$} & \multirow{2}{*}{$\checkmark$}  & \multirow{2}{*}{$\checkmark$} & & Baseline & 55.4 & 62.7 & 60.7 & 55.0 & \multirow{2}{*}{\textcolor{red}{\bf $\uparrow$4.12}}\\
& & & & & \red{\textbf{+AMCL}} & \textbf{58.5} & \textbf{66.3} & \textbf{67.0} & \textbf{58.4} & \\
\hline
\multirow{2}{*}{$\checkmark$} & \multirow{2}{*}{$\checkmark$} & \multirow{2}{*}{$\checkmark$}  & \multirow{2}{*}{$\checkmark$} & \multirow{2}{*}{$\checkmark$} & Baseline & 57.6 & 64.4 & 63.6 & 58.2 & \multirow{2}{*}{\textcolor{red}{\bf $\uparrow$4.59}}\\
& & & & & \red{\textbf{+AMCL}} & \textbf{61.8} & \textbf{69.3} & \textbf{68.9} & \textbf{62.1} & \\
\bottomrule
\end{tabular}
\end{center}
\vspace{-0.5cm}
\end{table}

\noindent{\bf Impact of the number of data augmentation types.} 
In Table~\ref{cifar100-aug}, under only 1-2 augmentations during SSL pretraining, the improvements in linear probing over the baselines are around 1\%.  When we further increase the number of augmentations, linear probing performance improves; importantly, AMCL becomes more and more useful: average improvement becomes as large as 4.59\% when five types of data augmentation are used. This suggests the existence of multiple similarity relations when many data augmentations are applied, validating our motivation and method design. 

\begin{table}[tb]
    \centering
    \caption{COCO object detection and segmentation results reveal two notable findings: (i) Mask Average Precision (AP) exhibits a similar trend to box AP, and (ii) the learned representations via AMCL demonstrate generalizability to other tasks.}
    \vspace{-0.2cm}
    \begin{tabular}{lccccccccc}
        \toprule
        &  \multicolumn{4}{c}{AP$^\text{box}$} & & \multicolumn{4}{c}{AP$^\text{mask}$} \\
        \cline{2-5}
        \cline{7-10}
        & \multicolumn{2}{c}{\bkbone{ViT-B}} & \multicolumn{2}{c}{\bkbone{ViT-L}} &&  \multicolumn{2}{c}{\bkbone{ViT-B}}& \multicolumn{2}{c}{\bkbone{ViT-L}}\\
        \midrule
        MoCo& 48.1 & & 49.2 & &&  43.0&  & 43.8 & \\
        MoCo\red{\textbf{+AMCL}}& \textbf{50.3} & {\textcolor{red}{\bf $\uparrow$2.2}} & \textbf{53.3} & {\textcolor{red}{\bf $\uparrow$4.1}} & & \textbf{45.1} &{\textcolor{red}{\bf $\uparrow$2.1}}& \textbf{47.0} & {\textcolor{red}{\bf $\uparrow$3.2}}\\
        LGP& 52.5 && 54.9 & & & 46.1 && 48.1&\\
        LGP\red{\textbf{+AMCL}}& \textbf{54.1} & {\textcolor{red}{\bf $\uparrow$1.6}}& \textbf{57.0} & {\textcolor{red}{\bf $\uparrow$2.1}}& & \textbf{48.3} &{\textcolor{red}{\bf $\uparrow$2.2}}& \textbf{51.1} &{\textcolor{red}{\bf $\uparrow$3.0}}\\
        \bottomrule
    \end{tabular}
    \label{tab:det-seg}
    \vspace{-0.3cm}
\end{table}

\noindent{\bf Generalizations to other tasks.} We provide results on COCO object detection and segmentation. Follow~\cite{he2022masked}, we finetune Mask R-CNN end-to-end on COCO~\cite{lin2014microsoft}, and the ViT backbones (ImageNet pretrained) are adapted to use with FPN~\cite{lin2017feature}. We report box AP and mask AP for objection detection and instance segmentation, respectively, in Table~\ref{tab:det-seg}. These results demonstrate that the learned representations are generalizable (see also Fig. \ref{fig:images}(c) and (d)).

\begin{table}[tb]
    \centering
    \caption{Computational cost analysis. We use R50 for SimCLR and MoCo, and ViT-B for CAN and LGP. The additional computational cost from ACML is around 5\%.}
    \vspace{-0.2cm}
    \resizebox{\linewidth}{!}{\begin{tabular}{lrrrr}
        \toprule
         & SimCLR (\red{\textbf{+AMCL}}) &  MoCo (\red{\textbf{+AMCL}}) & CAN (\red{\textbf{+AMCL}}) & LGP (\red{\textbf{+AMCL}})\\
        \midrule
        \#Params(M)& 25.6 (\red{\textbf{+1.1}}) & 25.6 (\red{\textbf{+1.1}}) &  87.0 (\red{\textbf{+2.1}}) & 194.3 (\red{\textbf{+2.1}})\\
        GFLOPs& 3.5 (\red{\textbf{+0.3}})  & 4.1 (\red{\textbf{+0.4}}) & 68.5 (\red{\textbf{+4.1}}) & 76.1 (\red{\textbf{+4.4}})\\
        \bottomrule
    \end{tabular}}
    \label{tab:cost}
    \vspace{-0.3cm}
\end{table}

\noindent{\bf Computational cost analysis.} We provide a comparison of the number of parameters and FLOPs between baselines and ACML. We choose R50 for SimCLR and MoCo, and ViT-B for CAN and LGP. Table~\ref{tab:cost} summaries the results. The additional computational cost from ACML is around 5\%, which is marginal. 

\section{Conclusion}

We address the challenge posed by complex pair similarity distributions under multiple augmentation types by introducing Adaptive Multi-Head Contrastive Learning (AMCL). AMCL utilizes multiple projection heads, each generating distinct features, along with a pair-wise adaptive temperature scheme. We derive the loss function, revealing insights into the relationship between the variance of pair distance distribution and temperature, as well as the physical meanings of the regularization term. Our method effectively separates pair similarity distributions. AMCL demonstrates experimental improvements in popular SSL methods across various backbones, numbers of labeled samples for linear probing, and augmentation types. Particularly beneficial under multiple augmentation types, AMCL aligns with our motivation.



%
%
\bibliographystyle{splncs04}
\bibliography{main}

\appendix
\title{
Adaptive Multi-head Contrastive Learning
--Appendix--} 

\titlerunning{Adaptive Multi-head Contrastive Learning}

\author{Lei Wang\thanks{Corresponding author.}\inst{, 1,2}\orcidlink{0000-0002-8600-7099}\and Piotr Koniusz\inst{2,1}\orcidlink{0000-0002-6340-5289} \and Tom Gedeon\inst{3}\orcidlink{0000-0001-8356-4909} \and Liang Zheng\inst{1}\orcidlink{0000-0002-1464-9500}\\
}

\institute{$^{1}$Australian National University \;
$^{2}$Data61/CSIRO \;
$^{3}$Curtin University \\
\email{\{lei.w, liang.zheng\}@anu.edu.au}, \email{piotr.koniusz@data61.csiro.au}, \email{tom.gedeon@curtin.edu.au}}

\maketitle

\setcounter{page}{1}
\setcounter{equation}{5}
\setcounter{table}{8}
\setcounter{figure}{7}
\section{Deriving the AMCL loss from MLE}
\label{sec:deriv}

This section details the derivation of our loss function based 
 on the maximum likelihood estimation (MLE) over
head-wise posterior distributions of positive samples given observations. We show that our derivation is connected to an m-estimator  \cite{Huber.Wiley}  whose log-likelihood employs Normal distributions \aka  Welsch functions that are known to model the observation noise via the heteroscedastic aleatoric uncertainty \cite{uncertainty1,uncertainty5,uncertainty4}.
Our adaptive temperature captures such an uncertainty. Tuning constant $\tau$ was shown before to help learn good contrastive representations~\cite{chen2020simple,he2020momentum}.
\cite{wang2021understanding} also demonstrated that temperature $\tau$ controls the strength of penalties on the hard negative samples and established its relationship with uniformity, illustrating that a well-chosen $\tau$ can strike a balance between the alignment and uniformity properties of contrastive loss.
\cite{kukleva2023temperature} has shown that in place of constant temperature, a cosine schedule can improve learning--a seemingly minor modification with large impact on the learned embedding space. 
For $\ell_2$ normalized vectors, the relationship between squared Euclidean distance $\lVert\cdot\rVert_2^2$ and cosine similarity measure is: $\lVert \vz_i \!- \!\vz_j \rVert_2^2 \!=\!2 \!- \!\text{sim}(\vz_i, \vz_j)$. 
The Normal distribution $\mathcal{N}$ relies on the squared Euclidean distance. 
To derive our multi-head NT-Xent loss, consider the following maximum likelihood estimation \textit{\wrt} parameters given as $\mathcal{P}=\big\{\boldsymbol{\theta},\{\tau^{c+}_i\}_{c=1}^C,\{\{\tau^{c-}_{in}\}_{n=1}^N\}_{c=1}^C\big\}$ and $\beta=1$:
\begin{align}
\label{eq:pr1}
\text{\fontsize{8}{8}\selectfont$\mathcal{P}^*$}&\text{\fontsize{8}{8}\selectfont$=\argmax\limits_{\mathcal{P}}  \prod_{c=1}^C \frac{\mathcal{N}\big(2-2\text{sim}(\vz^c_i, \vz^{c+}_i); \tau^{c+}_i \big)}{\sum_{n=1}^N \mathcal{N}\big(2-2\text{sim}(\vz^c_i, \vz^{c-}_{in}); \tau^{c-}_{in} \big)}$}\\
&\text{\fontsize{8}{8}\selectfont$=\argmin\limits_{\mathcal{P}} \sum_{c=1}^C\Big(\!\!-\log\mathcal{N}\big(2-2\text{sim}(\vz^c_i, \vz^{c+}_i); \tau^{c+}_i \big)+ \log\sum_{n=1}^N \mathcal{N}\big(2-2\text{sim}(\vz^c_i, \vz^{c-}_{in}); \tau^{c-}_{in} \big)\Big)$}\nonumber\\
&\text{\fontsize{8}{8}\selectfont$=\argmin\limits_{\mathcal{P}}\sum\limits_{c=1}^C\!\Big(\! \!-\frac{1}{\tau^{c+}_i}\text{sim}(\vz^c_i, \vz^{c+}_i) + \beta\Omega(\tau^{c+}_i) $}\nonumber\\
& \text{\fontsize{8}{8}\selectfont$+\log\sum_{n=1}^N\frac{1}{(2\pi)^{d'/2}(\tau^{c-}_{in})^{d'/2}}\exp\Big(\frac{1}{\tau^{c-}_{in}}\big(\text{sim}(\vz^c_i, \vz^{c-}_{in})-1\big)\Big)$}.
\label{eq:jump}
\end{align}
In Eq. \eqref{eq:jump}, we simply use expansion: 
\begin{equation}
\text{\fontsize{8}{8}\selectfont$-\log\bigg(\frac{1}{(2\pi)^{{d'}/{2}}(\sigma^2)^{d'/{2}}}\exp\Big(-\frac{2-2\mathbf{s}}{2\sigma^2} \Big)\bigg)=d'/2\log(2\pi)+(d'/2)\log(\sigma^2)+1/\sigma^2-\mathbf{s}/\sigma^2$,}
\end{equation}
where variance $\sigma^2=\tau$. We drop the constant (no impact on optimization) and are left with $-\mathbf{s}/\tau$ and $\Omega(\tau)=(d'/2)\log(\tau)+1/\tau$. 
We apply approximation in Eq. \eqref{eq:approx1} to Eq. \eqref{eq:jump} (rightmost part) and readily obtain Eq. \eqref{eq:ntxent2}. To derive multi-head InfoNCE loss, we solve a slightly modified problem:
\begin{align}
\label{eq:jump2}
\text{\fontsize{8}{8}\selectfont$\mathcal{P}^*$}&\text{\fontsize{8}{8}\selectfont$=\argmax\limits_{\mathcal{P}}  \prod_{c=1}^C \frac{\mathcal{N}\big(2-2\text{sim}(\vz^c_i, \vz^{c+}_i); \tau^{c+}_i \big)}{\mathcal{N}\big(2-2\text{sim}(\vz^c_i, \vz^{c+}_i); \tau^{c+}_i \big)+\sum_{n=1}^N \mathcal{N}\big(2-2\text{sim}(\vz^c_i, \vz^{c-}_{in}); \tau^{c-}_{in} \big)}$}\\
&\text{\fontsize{8}{8}\selectfont$=\argmax\limits_{\mathcal{P}}  \prod_{c=1}^C \frac{\mathcal{N}\big(2-2\text{sim}(\vz^c_i, \vz^{c+}_i); \tau^{c+}_i \big)}{\sum_{n=1}^{N+1} \mathcal{N}\big(2-2\text{sim}(\vz^c_i, \vz^{c\pm}_{in}); \tau^{c\pm}_{in} \big)}$}
\text{\fontsize{8}{8}\selectfont$=\argmax\limits_{\mathcal{P}}  \prod_{c=1}^C p(\vz^{c+}_{ic}|\vz^c_i)$} \nonumber \\
& \text{\fontsize{8}{8}\selectfont$=\argmax\limits_{\mathcal{P}}  \prod_{c=1}^C\frac{p(\vz^c_i|\vz^{c+}_{ic})p(\vz^{c+}_{ic})}{p(\vz^{c}_{i})}$},\nonumber
\end{align}
where $p(\vz^{c}_{i})\!=\!\sum_{n=1}^{N+1} \mathcal{N}\big(2-2\text{sim}(\vz^c_i, \vz^{c\pm}_{in}); \tau^{c\pm}_{in} \big)$, $p(\vz^{c+}_{ic})$ is a constant, \eg, 1,  and $p(\vz^c_i|\vz^{c+}_{ic})=\mathcal{N}\big(2-2\text{sim}(\vz^c_i, \vz^{c+}_i); \tau^{c+}_i \big)$.
Thus, the ratio of Gaussians in Eq. \eqref{eq:jump2} can be interpreted as maximizing head-wise posterior distributions of positive samples given observations.


\noindent
\textbf{Connecting temperature to uncertainty.}
Eq.~\eqref{eq:pr1} uses the variance $\tau$ of the distribution of pair-wise distances. Eq.~\eqref{eq:pr1} derives Eq.~\eqref{eq:jump}, where $\tau$ weighs the similarity, making it effectively the temperature. Because variance is usually treated as uncertainty~\cite{zhang2021temperature, wang2022uncertainty}, we build natural correspondence between uncertainty and temperature.
As we derive our multi-head losses (\eg, InfoNCE) from the MLE, we optimize this problem over network parameters and temperature (parametrized by an MLP). The temperature is tied with Welsch functions (Gaussians) in Eq. \eqref{eq:pr1} and \eqref{eq:jump2} whose radii are known to determine their influence range (tolerance to outliers).

\section{More Discussions}
\label{app:discussion}

\begin{figure}[tbp]
    \centering
    \includegraphics[trim=0cm 9.5cm 4.3cm 1.8cm, clip=true, width=\linewidth]{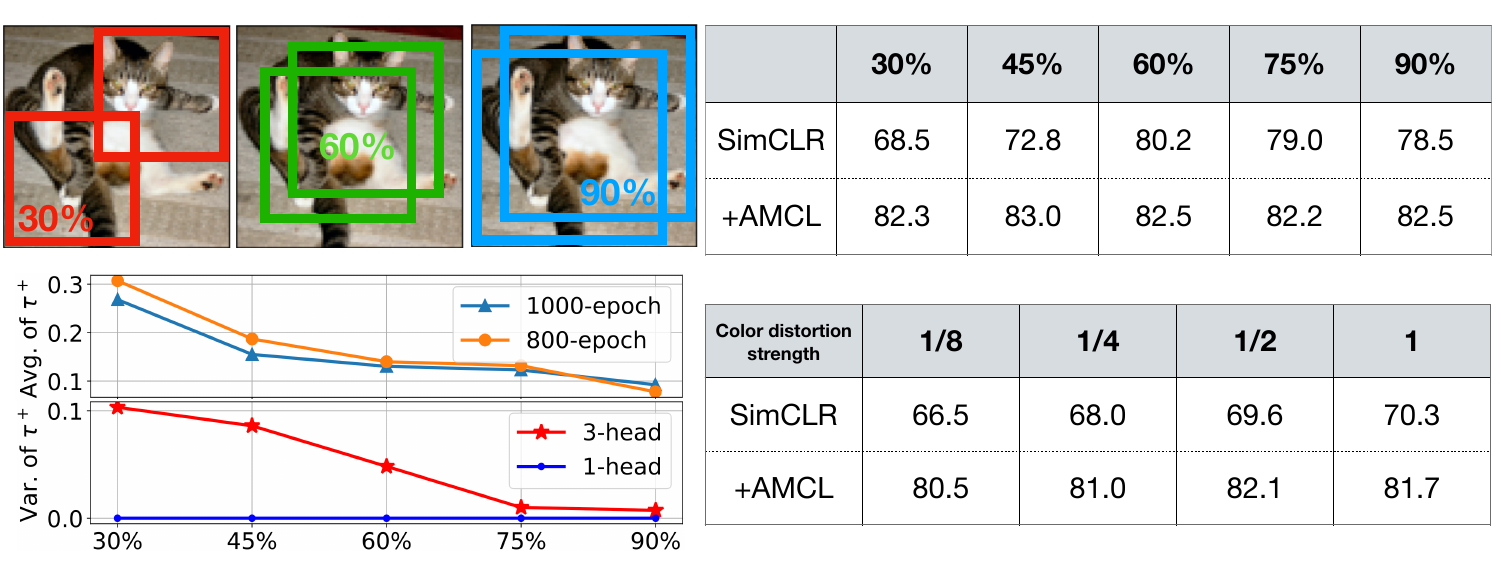}
    \caption{(\textit{Top left}) Overlap percentages of crops between positive pairs. (\textit{Top right}) Evaluations of the effects of overlap percentages. (\textit{Bottom left}) Average and variance of temperatures of positive pairs with different overlap percentages; the red curve represents the average sample-wise variance of temperatures from three heads. (\textit{Bottom right}) Evaluation of different color distortion strengths.}
    \label{fig:supp-rebuttal}
\end{figure}

\noindent\textbf{Criterion to define positive pair.}
Positive pairs that form two views are generated by several augmentations of an image. Fig.~\ref{fig:motivation} (pair indicated by green dot) in the main paper shows different crops of a sheep (1st pair) and car colors/shapes (4th pair). For stronger augmentations (\eg, low overlap of two crops) the noise effect on the contrastive loss becomes stronger (\eg, disjoint positive box of cat's leg may be shared between different heads). Thus, our positive-pair temperature obtained from the aleatoric uncertainty learner downweights particularly difficult noisy positive pairs but is penalized by $\Omega$ to avoid unnecessary downweighting. Fig.~\ref{fig:supp-rebuttal} (\textit{top right}) shows that for small overlaps (\eg, 30\% red box in Fig.~\ref{fig:supp-rebuttal} (\textit{top left})), our SimCLR+AMCL recovers performance, while SimCLR performs poorly. Fig.~\ref{fig:supp-rebuttal} (\textit{bottom right}) shows the same trend for heavy color distortion.
Fig.~\ref{fig:supp-rebuttal} (\textit{bottom left}) shows the average  (over epochs) temperature of positive-pair temperatures is high if crops overlap 30\%, indicting high uncertainty. For 90\% overlap, uncertainty  drops. 

\noindent\textbf{Why not use multi-head intrinsic features consistent across different heads?} This approach is handcrafted. Instead, we allow each head to specialize driven by the data, similar to multiple attention heads in a transformer. As each head is initialized differently, it captures various aspects of the data. Fig.~\ref{fig:supp-rebuttal} (\textit{bottom left}) red curve shows the average sample-wise variance of temperatures from three heads. High variances indicate that the temperature of each head differs, so each head's alignment varies (global/local for high/low temperature). In  experiments, a single head could not efficiently capture different aspects of the content. In contrast, multi-head captures complementary aspects of similarity between views, \eg, attributes, textures, shapes, \etc, due to a pair-adaptive head-wise temperature (Fig.~\ref{fig:motivation} (b)-(g) in the main paper), contributing to a more robust and refined similarity measure (Fig.~\ref{fig:images} in the main paper).

\noindent\textbf{Why did this method outperform SOTA?} Our adaptive temperature is based on aleatoric uncertainty modeling, which adapts heads to difficult positive/negative pairs.

\noindent\textbf{Why not use multiple backbones to improve feature learning?} This idea has been explored in supervised learning~\cite{tao2019deep}. However, training multiple backbones imposes prohibitive computational costs in SSL with no guarantee of the complementarity of such backbones.

\noindent\textbf{Adaptive temperature \vs attention learning.} The latter assigns varying weights to different components or parts of an object according to a specific design \cite{DBLP:journals/corr/BahdanauCB14,chorowski2015attention,caron2021emerging}. 
The learnable positive and negative temperatures reweigh the similarities by considering diverse image content resulting from multiple augmentations. This correction replaces the global temperature, allowing the backbone and multiple projection heads to focus on capturing different aspects of image content. Moreover, pair-wise weighted similarities on `alignment' and `uniformity' 
allow  various similarity relations to contribute differently to contrastive learning, similar to an attention learning mechanism.

\begin{figure}[tbp]
    \centering
    \includegraphics[trim=1.5cm 0cm 0cm 0cm, width=0.8\linewidth]{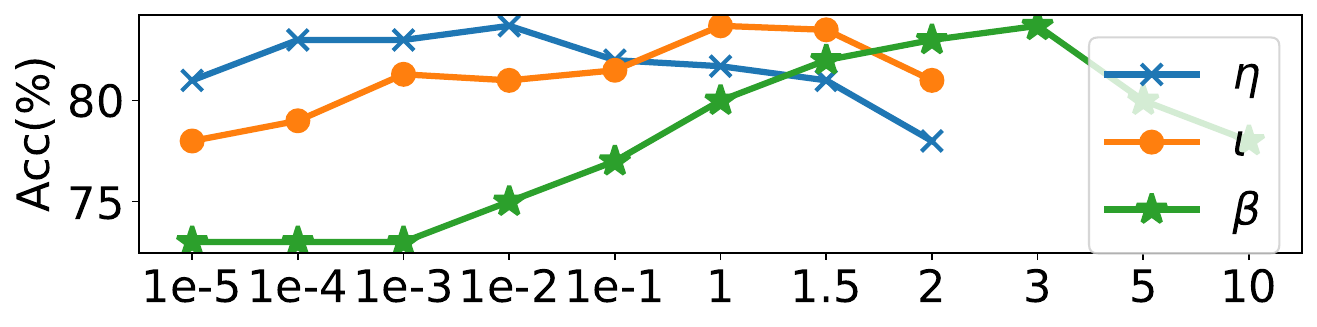}
    \caption{Sensitivity analysis of $\eta$, $\iota$, and $\beta$ on STL-10.}
    \label{fig:app-params-analysis}
\end{figure}

\noindent\textbf{Sensitivity analysis of $\eta$, $\iota$, and $\beta$.} We use Hyperopt package for hyperparameter optimization, running a total of 25 iterations. The search spaces for $\eta$, $\iota$, and $\beta$ are $[1e\!-\!5, 2]$, $[1e\!-\!5, 2]$, and $[1e\!-\!5, 10]$, respectively, as mentioned in the main paper. Fig.~\ref{fig:app-params-analysis} shows the sensitivity analysis of $\eta$, $\iota$, and $\beta$ on the STL-10 dataset.

\section{Dataset details}
\label{app:data}

We choose popular datasets that are widely used in evaluating the SSL models.

\noindent{\bf CIFAR-10}~\cite{Krizhevsky2009LearningML} consists of 60,000 $32\!\times\!32$ colour images divided into 10 classes, each containing 6,000 images. The dataset is split into 50,000 training images and 10,000 test images.

\noindent{\bf CIFAR-100} is similar to CIFAR-10 but comprises 100 classes, each with 600 images. There are 500 training images and 100 testing images per class. The 100 classes in CIFAR-100~\cite{Krizhevsky2009LearningML} are grouped into 20 superclasses. Each image is labeled with both a `fine' label (indicating its specific class) and a `coarse' label (indicating its superclass).

\noindent{\bf STL-10} 
~\cite{pmlr-v15-coates11a} is similarly to CIFAR-10 and includes images from 10 classes: airplane, bird, car, cat, deer, dog, horse, monkey, ship, truck. 
This dataset is relatively large and features a higher resolution ($96 \!\times\!96$ pixels) compared to CIFAR10.
It also provides a substantial set of $100,000$ unlabeled images that are similar to the training images but are sampled from a wider range of animals and vehicles. This makes the dataset ideal for showcasing the benefits of self-supervised learning. 

\noindent{\bf Tiny-ImageNet}~\cite{Le2015TinyIV} contains 100,000 images of 200 classes (500 for each class) downsized to $64\!\times\!64$ colored images. Each class has 500 training images, 50 validation images, and 50 test images.

\noindent{\bf ImageNet}~\cite{5206848} (\aka~{\bf ImageNet-1K}) contains 14,197,122 annotated images according to the WordNet hierarchy. Since 2010 the dataset is used in the ImageNet Large Scale Visual Recognition Challenge (ILSVRC), a benchmark in image classification and object detection. The publicly released dataset contains a set of manually annotated training images. 

\section{Impact statement}

This paper presents work whose goal is to advance the field of machine learning. There are many potential societal consequences of our work, none of which we feel must be specifically highlighted here.


\end{document}